\documentclass[lettersize,journal]{IEEEtran}
\usepackage{amsmath,amsfonts}
\usepackage{algorithmic}
\usepackage{algorithm}
\usepackage{array}
\usepackage[caption=false,font=normalsize,labelfont=sf,textfont=sf]{subfig}
\usepackage{textcomp}
\usepackage{stfloats}
\usepackage{url}
\usepackage{verbatim}
\usepackage{graphicx}
\usepackage{cite}

\usepackage{booktabs}
\usepackage{makecell} 
\usepackage{colortbl} 
\usepackage{xcolor}

\begin{document}

\title{Mitigate Target-level Insensitivity of Infrared Small Target Detection via Posterior Distribution Modeling}

\author{Haoqing Li, Jinfu Yang, Yifei Xu, Runshi Wang
\thanks{This work was supported in part by the National Natural Science Foundation of China under Grant no.61973009. \textit{(Corresponding author: Jinfu Yang.)}

Haoqing Li, Runshi Wang, and Yifei Xu are with the Faculty of Information, Beijing University of Technology, Beijing, 100124, P. R. China. (e-mails: lihaoqing@emails.bjut.edu.cn; wangrunshi@emails.bjut.edu.cn; xuyifei@emails.bjut.edu.cn). 

Jinfu Yang is with the Faculty of Information Technology, Beijing University of Technology, Beijing, 100124, China, also with the Beijing Key Laboratory of Computational Intelligence and Intelligent System, Beijing University of Technology, Beijing, 100124, P. R. China (e-mail: jfyang@bjut.edu.cn).}
\thanks{Manuscript received xxxx; revised xxxx.}}

\markboth{Journal of \LaTeX\ Class Files,~Vol.~14, No.~8, August~2021}%
{Shell \MakeLowercase{\textit{et al.}}: A Sample Article Using IEEEtran.cls for IEEE Journals}

\IEEEpubid{}

\maketitle

\begin{abstract}
Infrared Small Target Detection (IRSTD) aims to segment small targets from infrared clutter background. Existing methods mainly focus on discriminative approaches, i.e., a pixel-level front-background binary segmentation. Since infrared small targets are small and low signal-to-clutter ratio, empirical risk has few disturbances when a certain false alarm and missed detection exist, which seriously affect the further improvement of such methods. Motivated by the dense prediction generative methods, in this paper, we propose a diffusion model framework for Infrared Small Target Detection which compensates pixel-level discriminant with mask posterior distribution modeling. Furthermore, we design a Low-frequency Isolation in the wavelet domain to suppress the interference of intrinsic infrared noise on the diffusion noise estimation. This transition from the discriminative paradigm to generative one enables us to bypass the target-level insensitivity. Experiments show that the proposed method achieves competitive performance gains over state-of-the-art methods on NUAA-SIRST, IRSTD-1k, and NUDT-SIRST datasets. Code are available at https://github.com/Li-Haoqing/IRSTD-Diff.
\end{abstract}


\section{Introduction}
\label{sec1}
\IEEEPARstart{I}{nfrared} Small Target Detection (IRSTD), a technique for finding small targets from infrared clutter background, provides many potential applications in remote sense, public security, and other fields with strong anti-disturbance imaging capability \cite{lu2006detecting, zhao2022single, demosthenous2015infrared}. However, infrared small targets tend to be smaller than 9 × 9 pixels, insufficiency of color and texture features, and consequently, submerged by clutter background. These constraints make IRSTD challenging and prone to false alarm and missing detection.
\begin{figure}[!t]
	\centering
	\includegraphics[width=0.48\textwidth]{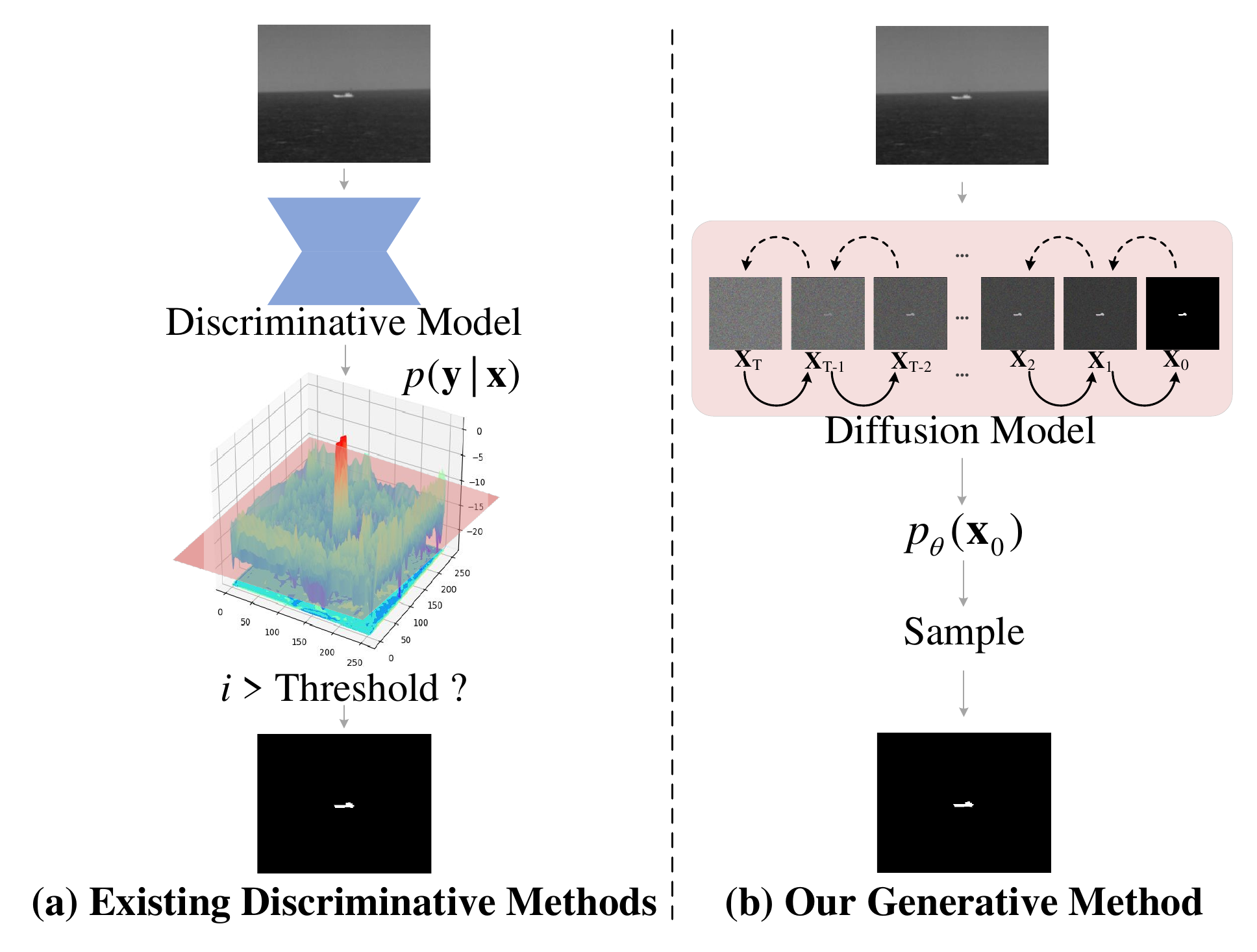}
	\caption{\textbf{Schematic comparison between (a) Existing Discriminative Methods and (b) Our Generative Method for Infrared Small Target Detection.} Our approach obtains the posterior distribution of small target masks.}
	\label{fig1}
\end{figure}
Conventional unlearnable methods, for instance, filtering-based \cite{deshpande1999max, Tophat}, local contrast-based \cite{TLLCM, WSLCM}, and low rank-based methods \cite{zhang2019infrared, sun2020infrared} are hindered by numerous hyper-parameters. Concurrently, their pixel-level performance is relatively inferior, with high false alarm and missing detection. In addition, deep learning methods \cite{ALC, uiunet} have achieved better pixel-level accuracy and lower false alarm and miss detection. Dai et al. \cite{ACM} used discriminative deep learning earlier and made a pioneering breakthrough. Li et al. \cite{DNA} proposed a dense nested attention network to incorporate and exploit contextual information. Zhang et al. \cite{ISNet} emphasized that shape matters and incorporated shape reconstruction into IRSTD. At present, the mainstream deep learning IRSTD methods are per-pixel discriminative learning, with Intersection over Union (IoU) \cite{yu2016unitbox} or Binary Cross Entropy (BCE) loss as the cost function. For example, methodologies such as ACM \cite{ACM} and DNANet \cite{DNA} calculate the IoU between predicted masks and ground truth as empirical loss, whereas UIUNet \cite{uiunet} and ILNet \cite{ILNet} employ pixel-level BCE for optimizing model parameters. However, due to the minuscule size of infrared small targets, the ratio of target pixels to the overall infrared image is exceedingly small. Consequently, the cost function has no significant disturbances when a certain false alarm and missed detection exist, owing to the rare pixels in the targets. The inconspicuous error is manifested in the false alarm rate (the prevailing state-of-the-art methods typically fall within the order of $10^{-6}$). Especially in the later training period with declined low learning rate, this insensitivity of the empirical risk seriously affects the further improvement of such methods.

In this paper, the generative diffusion framework is employed to obtain the mask distribution directly, instead of pixel-level discrimination. The scattered false alarm pixels bring strong disturbances to the simple data pattern, but the KL divergence constraint on the mask posterior distribution can suppress such outliers. Therefore, provided conditions of input infrared image, we maximize the posterior distribution of the small target mask to surmount the performance bottleneck associated with minimizing discriminative empirical risk. The final detection results are obtained by sampling from this distribution, as shown in Figure \ref{fig1}.

However, there is inevitably a massive of noise in the infrared imaging process, including background and hardware noise. The global white noise under the influence of background radiation, as well as shot noise, thermal noise, and low-frequency 1 / f noise caused by hardware in the thermal imaging, leads to an extremely low signal-to-clutter ratio \cite{vollmer2021infrared, 2005Noise}. These interferences will affect the noise estimation of the diffusion model. Several prophase image processing approaches \cite{malfait1997wavelet, jansen1999multiple, strela1999application, weyrich1998wavelet} used Wavelet Transform \cite{haar}, i.e., transform-filter-inverse, to enhance the infrared image, which was beneficial to IRSTD and other tasks. Nevertheless, these complex approaches cannot be applied to end-to-end diffusion model training. Based on the basic concept of "the infrared background remains in the low-frequency component, while the targets with higher intensity are reflected in the high-frequency" \cite{li2020infrared}, a low-frequency isolation module in the wavelet domain is designed. The low-frequency component of the infrared image in the wavelet domain is processed by a neural network, and the high-frequency component is utilized to restore the targets information. Finally, a residual of the enhanced features is estimated.

The contributions of this paper can be summarized as:

\begin{itemize}
	\item We examine the loss function of infrared small target detection and highlight a potential issue in discriminative training, i.e., the target-level insensitivity.
	\item A diffusion model framework for Infrared Small Target Detection is proposed. We generatively obtain the posterior distribution of small target masks, to surmount the performance bottleneck associated with minimizing discriminative empirical risk.
	\item A Low-frequency Isolation module in the Wavelet domain is designed, to reduce the influence of low-level infrared interference on diffusion noise estimation.
	\item Our method is qualitatively and quantitatively evaluated on three datasets and show a better performance than the existing state-of-the-art discriminative methods.
\end{itemize}

\section{Related Work}

\textbf{Infrared Small Target Detection:} Infrared Small Target Detection is widely investigated using different approaches. These include unlearnable methods: filtering-based, local contrast-based, and low rank-based methods; and deep learning methods: ACM \cite{ACM}, DNA \cite{DNA}, ISNet \cite{ISNet}, etc. The data-driven deep learning methods have better accuracy and convenient inference, without numerous manual hyper-parameters. Dai et al. \cite{ACM} first introduced the data-driven deep learning method into infrared small target detection and proposed an asymmetric contextual modulation module. Furthermore, they combined discriminative networks and conventional model-driven methods to utilize the infrared images and domain knowledge \cite{ALC}. Zhang et al. \cite{AGPC} proposed an attention-guided pyramid context network and further improved the applicability of data-driven methods of IRSTD. Li et al. \cite{DNA} proposed a dense nested attention network and a channel-spatial attention module for adaptive feature fusion and enhancement, to incorporate and exploit contextual information. Zhang et al. \cite{ISNet} indicated that shape matters for IRSTD and proposed Taylor finite difference (TFD)-inspired edge block aggregates to enhance the comprehensive edge information from different levels. Our previous work \cite{ILNet} found that the infrared small target lost information in the deep layers of the network, and proved that the low-level feature matters for recovering the disappeared target information.
These data-driven deep learning methods are discriminative, which are prone to generate plenty of false alarm and miss detection in the process of pixel-level discrimination. This phenomenon caused by the target-level insensitivity during discriminative training. In this paper, we surmount this problem with generative posterior distribution modeling.

\textbf{Diffusion model:} Diffusion model \cite{DDPM}, a prevailing and promising generative model at present, is superior to other types of generative models such as GANs \cite{GAN1, GAN2} and VAEs \cite{VAE1, VAE2}. The Conditional Diffusion Probabilistic Models \cite{liu2023more, nichol2021glide} use class embedding to guide the image generation process to a certain extent. These methods have led to some excellent large-scale applications, e.g., Imagen \cite{saharia2022photorealistic}, ERNIE-ViLG \cite{feng2023ernie}, and Stable Diffusion \cite{rombach2022high}, which made significant contributions to AI-Generated Content.

The Diffusion model has been applied to many generative tasks, such as inpainting \cite{lugmayr2022repaint}, image translation \cite{zhao2022egsde}, super-resolution \cite{saharia2022image}, etc. Similarly, the \textbf{discriminative application} has also been explored. Xiang et al. \cite{xiang2023denoising} proved that the denoising diffusion autoencoders (DDAE) are unified self-supervised learners and the diffusion pre-training is a general approach for self-supervised generative and discriminative learning. Therefore, the generative pre-training paradigm supports the diffusion model as a feasible discriminative learning method. Baranchuk et al. \cite{baranchuk2021label} demonstrate that diffusion models can serve as an instrument for dense prediction tasks, especially in the setup when labeled data is scarce. Simultaneously, Amit et al. \cite{sigdiff} used the diffusion model for image semantic segmentation, Wu et al. \cite{medsegdiff, medsegdiff2} used it to address the medical image segmentation problem, and Ma et al. \cite{ma2023diffusionseg} demonstrate the superiority of adapting diffusion for unsupervised object discovery. These works, as the explorers applied the diffusion model to discriminative tasks, have proved the feasibility and excellent performance.

Additionally, Chen et al. \cite{chen2023generative} regarded semantic segmentation as an \textbf{image-conditioned mask generation}, which is replacing the conventional per-pixel discriminative learning with a latent prior learning process. Le et al. \cite{le2023maskdiff} models the underlying conditional distribution of a binary mask, which is conditioned on an object region and K-shot information. These methods provided a novel approach to pixel-level discriminative tasks such as IRSTD.

In this work, we replace pixel-level discriminant with mask posterior distribution modeling. A pertinence conditional diffusion model is employed to achieve this purpose.

\section{Method}

\subsection{Motivation}

\begin{figure}[t]
	\centering
	\includegraphics[width=0.48\textwidth]{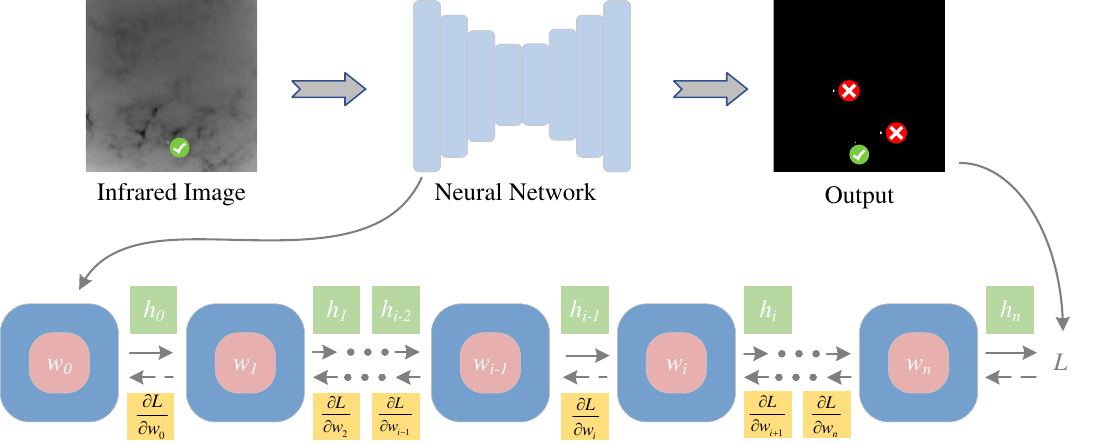}
	\caption{\textbf{Sketch map of the Back-propagation process.} The arrows to the right represent the Forward-propagation, while the arrows to the left represent the Back-propagation. For ease of representation, the bias and activation layers are omitted. $w_i$ and $h_i$ are the parameter and output of the $i$-th layer, respectively. All these partial derivatives in the IRSTD networks are exceedingly diminutive, but false alarms exist.}
	\label{fig3}
\end{figure}

During the experiments of IRSTD, we found that the empirical risk becomes quite low when the neural network is trained to the later epochs. But there are still some target-level errors, and these false alarm and miss detection unable to be eliminated. What causes this phenomenon?

We conducted extensive research on the loss function and back-propagation of the IRSTD model \cite{ACM, DNA, uiunet, ILNet}. Our research reveals that the IoU loss demonstrates insensitivity to false alarms, but sensitivity to miss detection. BCE loss exhibits insensitivity to all the target-level errors. IoU loss is computed as the average within a minibatch:
\begin{equation}
	L_{IoU}=1-\frac{1}{N}\sum_{i}^{N}\frac{TP}{T+P-TP},
\end{equation}
where $N$ is the batch size, $T$, $P$, and $TP$ denote the predicted, ground truth, and true positive pixels, respectively.
For instance, in a batch of size 8, if there are 9 false alarm pixels, the loss is mere 0.0103, whereas the loss for a miss detection in a single sample amounts to 0.125 (assuming a resolution of 256 × 256). In addition, the BCE loss is more insensitive, yielding a loss of only 8 × $10^{-5}$ for the same false alarm scenario. Discriminative empirical risk is progressively insensitive as the resolution increases.

The insensitivity exerts a significant influence on the parameter updates, allowing these error phenomena unpunished. Illustrated in Figure 2, the minimal empirical risk results in all pertinent partial derivatives $\frac{\partial L}{\partial w_i} (i=0,1,…,n)$ exceedingly diminutive. Coupled with the learning rate decay to a small value in the later epochs, the optimal parameters are struggling to obtain. Our methodology amalgamates generative paradigm to address this performance bottleneck.

\subsection{Background}

Diffusion model \cite{DDPM}, a type of generative latent variable model, includes \textit{forward} and \textit{reverse process}. The \textbf{forward process} adds Gaussian noise to the original image $\mathbf{x_0}$ according to a given variance table $[\beta_1, ..., \beta_T]$:
\begin{equation}
	q(\mathbf{x}_t| \mathbf{x}_{t-1}):=\mathcal{N}(\mathbf{x}_t;\sqrt{1-\beta_t}\mathbf{x}_{t-1},\beta_t\mathbf{I}).
\end{equation}
The Markov process lasts \textit{T} steps until the approximate standard Gaussian noise $\mathcal{N}(\mathbf{x}_T;\mathbf{0},\mathbf{I})$ is obtained:
\begin{equation}
	q(\mathbf{x}_{1:T}| \mathbf{x}_{t-1}):=\sum_{t=1}^{T}q(\mathbf{x}_{t}|\mathbf{x}_{t-1}),
\end{equation}
where $[\mathbf{x}_1, ..., \mathbf{x}_t, ... \mathbf{x}_T]$ is the implicit variable of the intermediate forward process, which can be obtained directly:
\begin{equation}
	q(\mathbf{x}_t| \mathbf{x}_{0}):=\mathcal{N}(\mathbf{x}_t;\sqrt{\bar{\alpha}_t}\mathbf{x}_{0},(1-\alpha_t)\mathbf{I}),
\end{equation}
where $\alpha_t:=1-\beta_t$ and $\bar{\alpha}_t:=\prod_{s=1}^{t}\alpha_s$.

The \textbf{reverse process} starts with a standard Gaussian noise $\mathbf{x}_T$, completes by \textit{T} steps learnable Gaussian transformation, and obtains the data distribution $p_\theta(\mathbf{x}_0)$ finally, which can be formulated as:
\begin{equation}
	p_\theta(\mathbf{x}_{t-1}| \mathbf{x}_{t}):=\mathcal{N}(\mathbf{x}_{t-1};\boldsymbol{\mu}_\theta(\mathbf{x}_t,t),\boldsymbol{\Sigma}_\theta(\mathbf{x}_t,t)),
\end{equation}
\begin{equation}
	\label{5}
	p_\theta(\mathbf{x}_{0:T})=p(\mathbf{x}_T)\prod_{t=1}^{T}p_\theta(\mathbf{x}_{t-1}|\mathbf{x}_t),
\end{equation}
\begin{equation}
	\label{6}
	p_\theta(\mathbf{x}_0):=\int p_\theta(\mathbf{x}_{0:T}) d\mathbf{x}_{1:T},
\end{equation}
where $\boldsymbol{\mu}_\theta(\mathbf{x}_t,t)=\frac{1}{\sqrt{\alpha_t}}(\mathbf{x}_t-\frac{\beta_t}{\sqrt{1-\bar{\alpha}_t}}\boldsymbol{\epsilon}_\theta(\mathbf{x}_t,t))$ is a learnable parameterized expectation, and $\boldsymbol{\epsilon}_\theta(\mathbf{x}_t,t)$ is a parameterized noise.  An unlearnable variance $\boldsymbol{\Sigma}_\theta(\mathbf{x}_t,t)$ with the configuration of \cite{DDPM} is utilized.

Training is to optimize the variable boundary of negative log-likelihood $\mathbb{E}[-logp_\theta(\mathbf{x}_0)]$, and minimize the KL divergence between $p_\theta(\mathbf{x}_{t-1}|\mathbf{x}_t)$ and the forward process posterior $q(\mathbf{x}_{t-1}|\mathbf{x}_t,\mathbf{x}_0)$, which can be simplified as:
\begin{equation}
	\label{7}
	L=\sum_{t>1}D_{KL}(q(\mathbf{x}_{t-1}|\mathbf{x}_t,\mathbf{x}_0)||p_\theta(\mathbf{x}_{t-1}|\mathbf{x}_t)),
\end{equation}
where, $D_{KL}(.)$ represents the KL divergence,
\begin{equation}
	q(\mathbf{x}_{t-1}|\mathbf{x}_t,\mathbf{x}_0)=\mathcal{N}(\mathbf{x}_{t-1};\tilde{\boldsymbol{\mu}}_t(\mathbf{x}_t,\mathbf{x}_0),\tilde{\beta}_t\mathbf{I}),
\end{equation}
\begin{equation}
	\tilde{\boldsymbol{\mu}}_t(\mathbf{x}_t,\mathbf{x}_0):=\frac{\sqrt{\bar{\alpha}_{t-1}}\beta_t}{1-\bar{\alpha}_t}\mathbf{x}_0 + \frac{\sqrt{\alpha_t}(1-\bar{\alpha}_{t-1})}{1-\bar{\alpha}_t}\mathbf{x}_t,
\end{equation}
\begin{equation}
	\tilde{\beta}_t:=\frac{1-\bar{\alpha}_{t-1}}{1-\bar{\alpha}_t}\beta_t.
\end{equation}
If a neural network is used to predict noise $\epsilon_\theta$, the Formula \ref{7} can be equivalent to optimizing the conditional expectation
\begin{equation}
	L'=\mathbb{E}_{\mathbf{x}_0,\boldsymbol{\epsilon},t}[||\boldsymbol{\epsilon}-\boldsymbol{\epsilon}_\theta(\sqrt{\bar{\alpha}_t}\mathbf{x}_0 + \sqrt{1-\bar{\alpha}_t}\boldsymbol{\epsilon},t)||^2],
\end{equation}
where $\boldsymbol{\epsilon} \sim \mathcal{N}(0, \mathbf{I})$ is a standard Gaussian noise of forward process posteriors.

\begin{figure*}[t]
	\centering
	\includegraphics[width=1\textwidth]{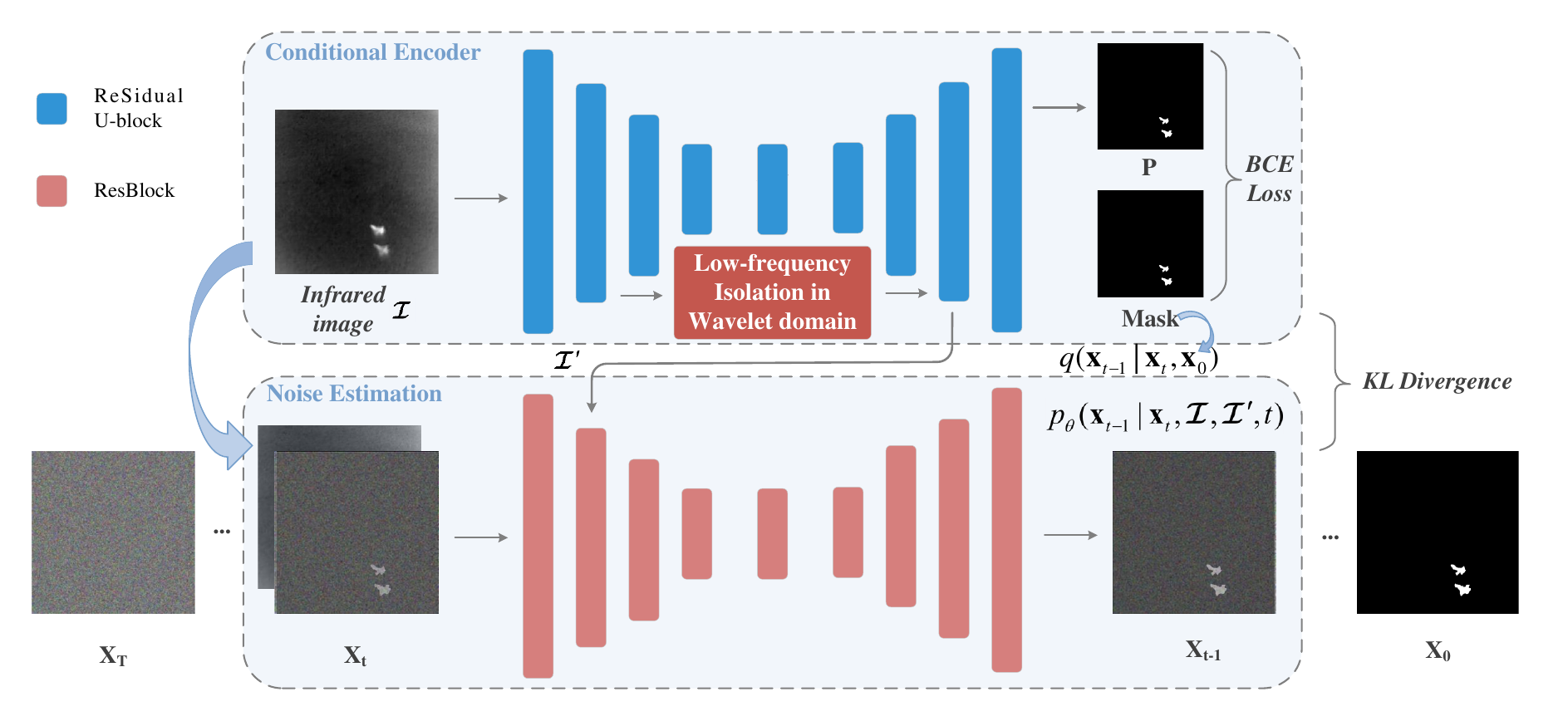} 
	\caption{\textbf{Overview of the proposed diffusion framework for Infrared Small Target Detection.} Each colored rectangle represents a corresponding Block, and the rectangle's size reflects the latent space level of the Blocks. The small gray arrows are in the same direction as the computational graph, while the large arrows are indicators.}
	\label{fig2}
\end{figure*}

\subsection{Overall Architecture}
A diffusion based approach, IRSTD-Diff, is proposed to obtain the underlying conditional mask distribution $p_\theta(\mathbf{x}_{t-1}| \mathbf{x}_{t},\boldsymbol{\mathcal{I}},t)$ for noised mask $\mathbf{x}_t$, conditioned on the infrared image $\boldsymbol{\mathcal{I}}$ and latent space encoding $\boldsymbol{\mathcal{I}}’$. Then, the posterior distribution of the original mask is obtained by the Markov process of Formula \ref{5} and \ref{6}. The conditional distribution $p_\theta(\mathbf{x}_{t-1}| \mathbf{x}_{t},\boldsymbol{\mathcal{I}},t)$ can be formulated as:
\begin{equation}
	\boldsymbol{\mu}_{\theta}(\mathbf{x}_{t},\boldsymbol{\mathcal{I}},\boldsymbol{\mathcal{I}}',t)=\frac{1}{\sqrt{\alpha_t}}(\mathbf{x}_t - \frac{\beta_t}{\sqrt{1-\bar{\alpha}_t}}\boldsymbol{\epsilon}_\theta(\mathbf{x}_{t},\boldsymbol{\mathcal{I}},\boldsymbol{\mathcal{I}}',t)),
\end{equation}
\begin{equation}
	p_{\theta}(\mathbf{x}_{t-1}|\mathbf{x}_{t},\boldsymbol{\mathcal{I}},\boldsymbol{\mathcal{I}}',t)=\mathcal{N}(\mathbf{x}_{t-1};\boldsymbol{\mu}_{\theta}(\mathbf{x}_{t},\boldsymbol{\mathcal{I}},\boldsymbol{\mathcal{I}}',t),\sqrt{\beta_t}\mathbf{I}),
\end{equation}
where, $\boldsymbol{\mu}_{\theta}$ is the parameterized expectation, $ \boldsymbol{\epsilon}_\theta(.)$ is a noise estimation neural network.

\subsection{Conditional Encoder}
As shown in Figure \ref{fig2}, the noise estimation network $ \boldsymbol{\epsilon}_\theta(.)$ is based on the original U-Net \cite{UNet} architecture of the DDPMs \cite{DDPM, IDDPM}and combined with the Conditional Encoder network (CE). CE, composed of ReSidual U-block \cite{U2Net}, is used to achieve the latent space perceptual compression of the infrared image $\boldsymbol{\mathcal{I}}$, and capture the semantic information of small targets. ReSidual U-block is a network with sub-UNets embedded in the UNet architecture. The nested U-shaped structure enables the network to capture abundant local and global information from shallow and deep layers, which is applicable to all resolutions. Previous studies \cite{uiunet, ILNet} have demonstrated that this multi-level nested Residual U-block as the backbone is more suitable for feature extraction of small infrared targets. Due to the high-level semantic information deficiency of small infrared targets, excessive latent space compression will remove the high-frequency target information. Therefore, we use the appropriate low-level latent space as the latent space embedding $\boldsymbol{\mathcal{I}}’$. $ \boldsymbol{\epsilon}_\theta(.)$ can be formulated as:
\begin{equation}
	\boldsymbol{\epsilon}_{\theta}(\mathbf{x}_{t},\boldsymbol{\mathcal{I}},\boldsymbol{\mathcal{I}}',t)=U_{de}(U_{en}(\mathbf{x}_t,t)+f(\boldsymbol{\mathcal{I}}),t),
\end{equation}
where, $U_{de}$ and $U_{en}$ denote U-Net encoder and decoder respectively, and $f(.)$ is the conditional encoder network.

The intricate architecture of our Conditional Encoder network (CE) is delineated in Table \ref{t2}. This encoder employs a nested sub-encoding and decoding structure, comprising six encoders and five decoders internally. The sub-encoder is employed to achieve the latent space perceptual compression of infrared images. Throughout this process, the low-level information of the infrared image is compressed, and the semantic structure is preserved for the targets reconstruction and restoration. This restoration reflects the necessity of the sub-decoder's existence, i.e., binary cross entropy discriminative training. Our generative architecture is not to completely abandon the discriminative paradigm, which is crucial for pixel-level infrared small target detection. In the absence of joint discriminative training, the sampling of the posterior distribution yield superior target-level performance, but struggle to achieve state-of-the-art pixel-level performance. This aspect will be thoroughly discussed in Section \ref{sec5}.

\textbf{Overall loss:}
To surmount the performance bottleneck of pixel-level discrimination, i.e., reduce false alarm and miss detection while maintaining pixel-level performance, we minimize the BCE-loss and the KL divergence between $p_\theta(\mathbf{x}_{t-1}|\mathbf{x}_t,\boldsymbol{\mathcal{I}},\boldsymbol{\mathcal{I}}',t)$ and posterior $q(\mathbf{x}_{t-1}|\mathbf{x}_t,\mathbf{x}_0)$ of the forward process:
\begin{equation}
	L=L' - \lambda[\mathbf{x}_0log(\mathbf{P})+(1-\mathbf{x}_0)log(1-\mathbf{P})],
\end{equation}
where, $\mathbf{P}$ is the estimated result of the conditional encoder, $\lambda$ is a factor to balance an optimal learning rate of two parts.

\begin{figure}[t]
	\centering
	\includegraphics[width=0.48\textwidth]{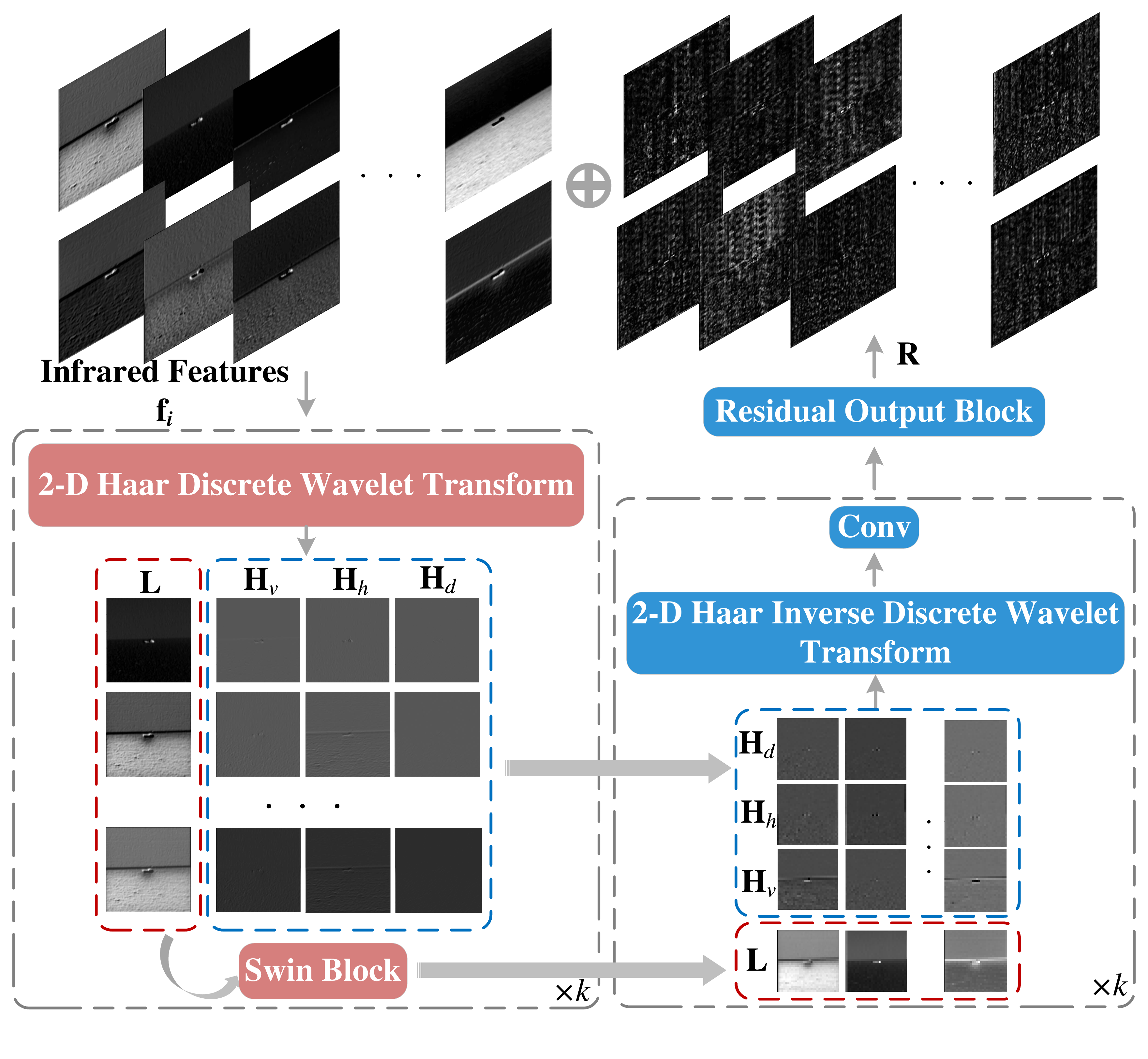}
	\caption{\textbf{Low-frequency Isolation in the Wavelet domain (LIW). L / H} represent the visualization of high / low-frequency components in the wavelet domain. \textbf{R} is the visualization of the estimated residuals.}
	\label{fig4}
\end{figure}

\begin{figure}[t]
	\centering
	\includegraphics[width=0.5\textwidth]{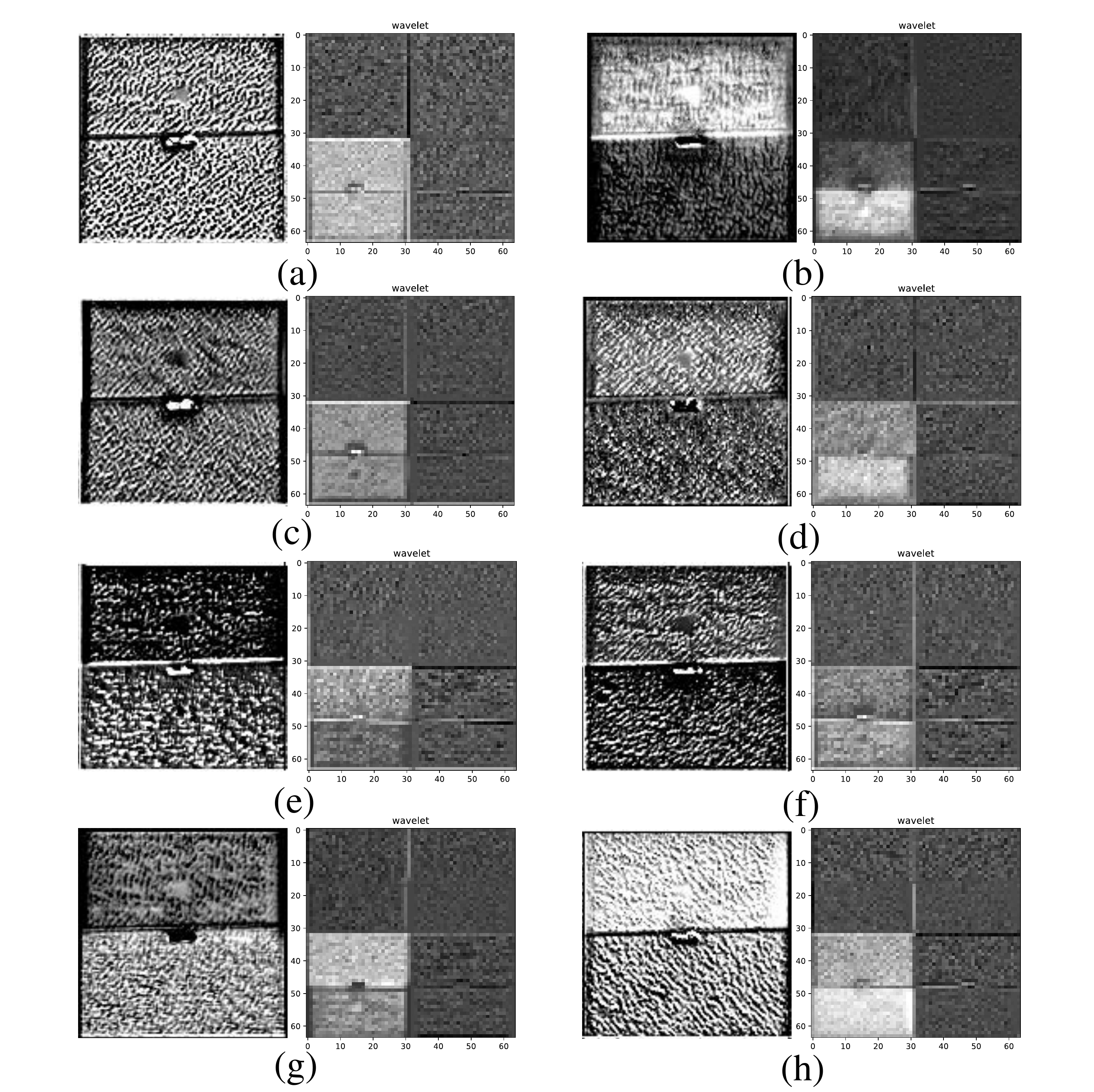}
	\caption{\textbf{One-level 2-D Haar Discrete Wavelet Transform (HDWT) \cite{haar} applied to the infrared features.} The left figures are Conditional Encoder output features, and the right figures are corresponding wavelet domain visualizations. The bottom left, bottom right, top left, and top right subfigures are low-frequency approximation and high-frequency horizontal, vertical, and diagonal components, respectively. These wavelet domain components are used as part of LIW.}
	\label{fig5}
\end{figure}

\begin{figure*}[t]
	\centering
	\includegraphics[width=1\textwidth]{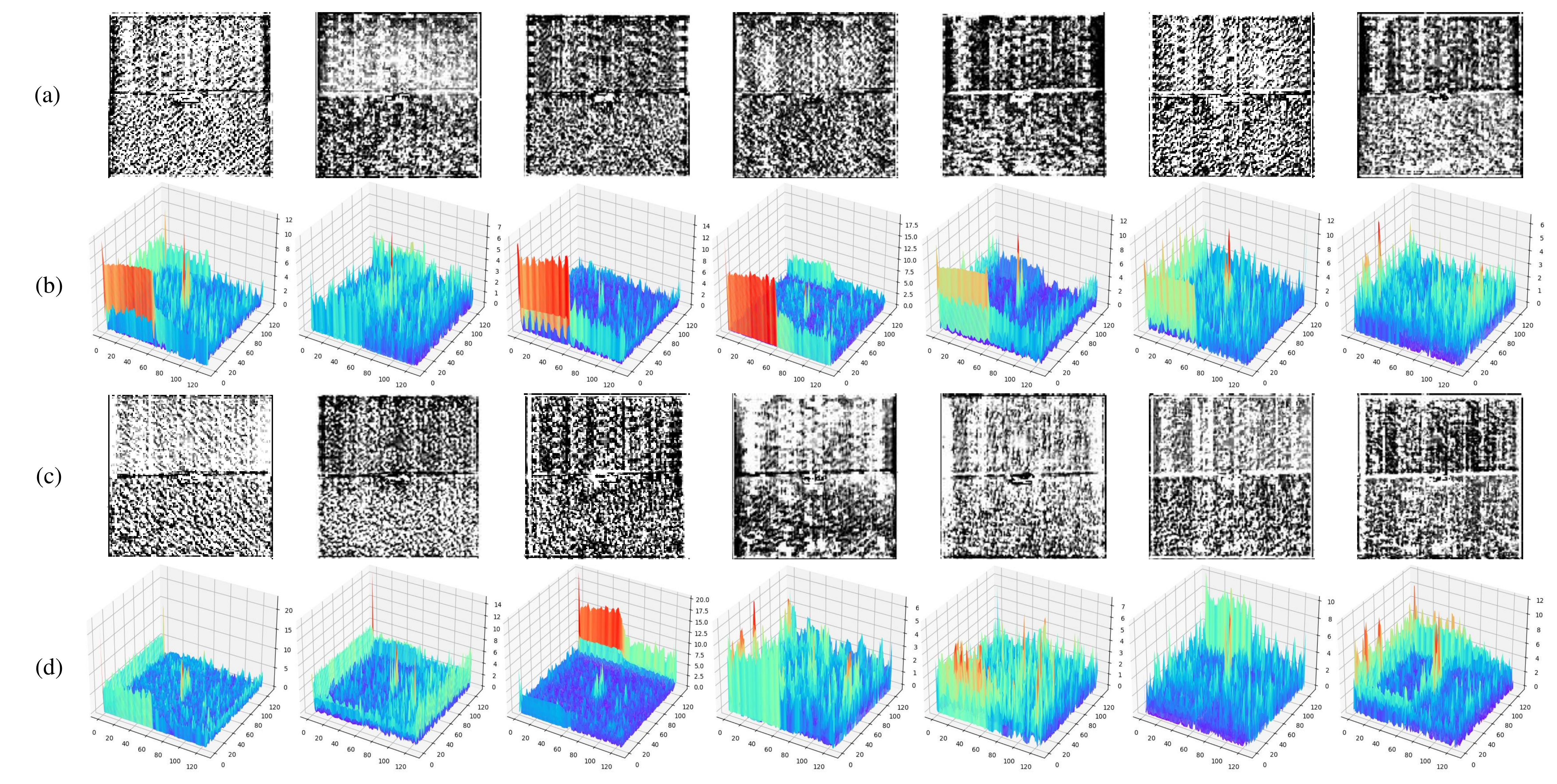}
	\caption{\textbf{The estimated residuals $\mathbf{R}$ and its 3D visualization, correspond to the infrared features in LIW.} (a) (b) are estimated residuals $\mathbf{R}$ of the partial channels, and (b) (d) are corresponding 3D visualization.}
	\label{fig6}
\end{figure*}

\begin{figure*}[t]
	\centering
	\includegraphics[width=1\textwidth]{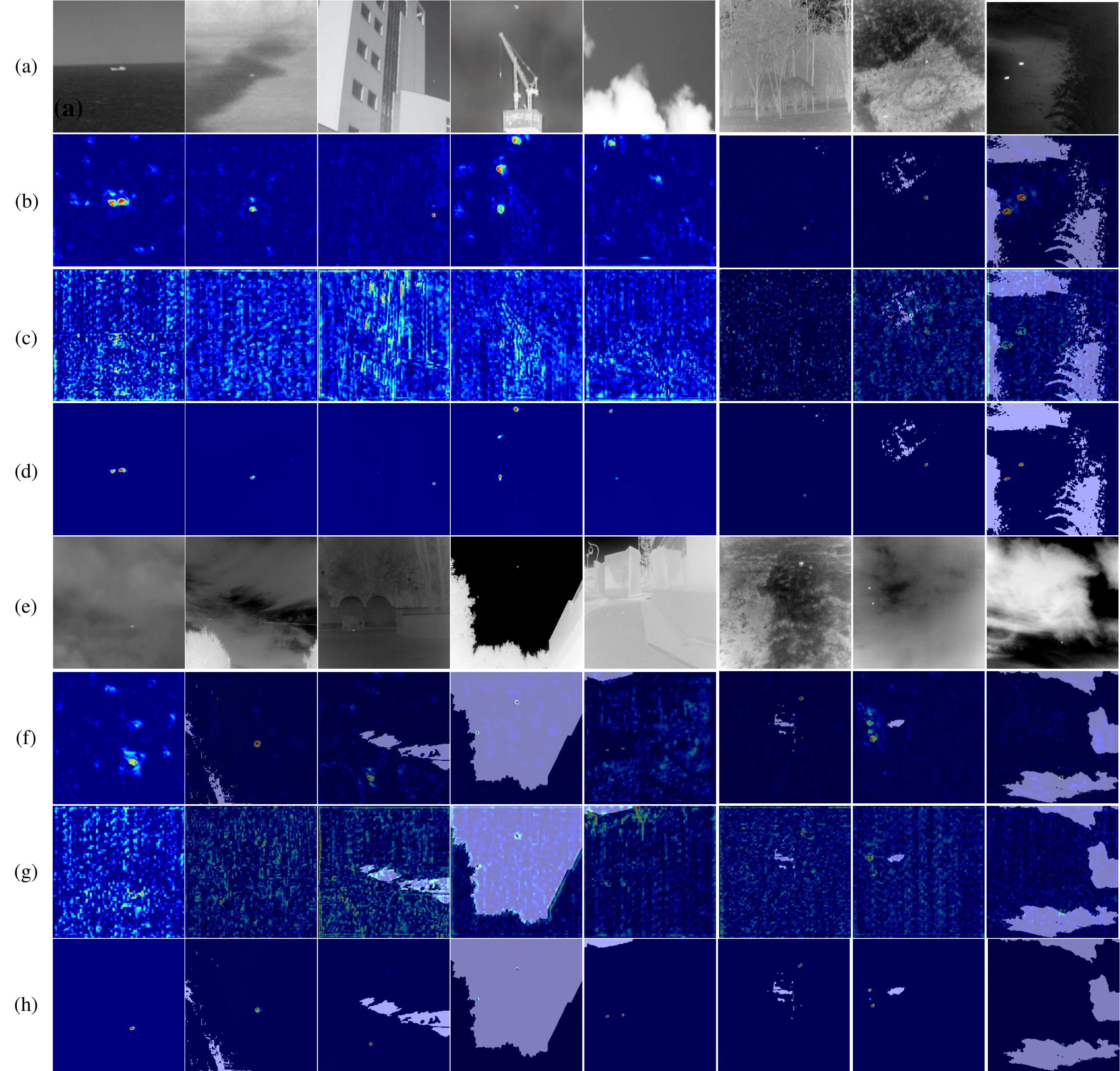}
	\caption{\textbf{Class Activation Mapping (CAM) comparison between (d) (h) our IRSTD-Diff and (b) (f) our model without LIW.} (a) (e) are input infrared images. LIW conspicuously isolates the background radiation and obtains a cleaner infrared conditional encoder. (c) (g) are residuals $\mathbf{R}$ estimated by LIW for isolating low-frequency disturbances.}
	\label{fig7}
\end{figure*}

\begin{figure}[t]
	\centering
	\includegraphics[width=0.5\textwidth]{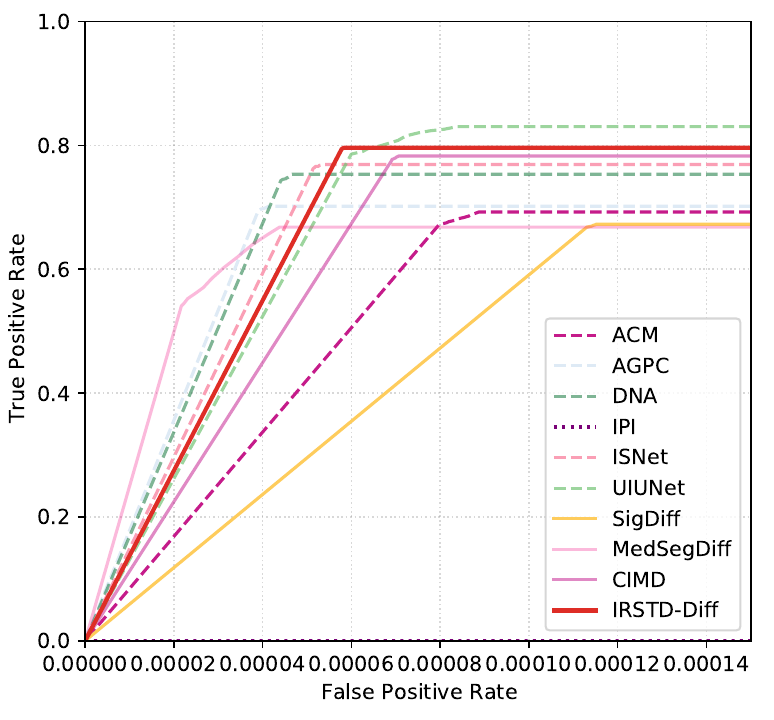}
	\caption{\textbf{Receiver Operating Characteristic (ROC)
			curves performance on the IRSTD-1k dataset.} The performance of our IRSTD-Diff is stable and better than other diffusion-based methods. It also has competitiveness compared with the discriminant methods.}
	\label{fig8}
\end{figure}

\subsection{Low-frequency isolation in the wavelet domain}
During the imaging process, infrared images inevitably exhibit substantial noise, leading to an exceedingly low signal-to-noise ratio. It is inevitable to introduce unnecessary interference when using low-level latent encoding as $\boldsymbol{\mathcal{I}}'$, owing to the particularity of small infrared targets. Therefore, a Low-frequency Isolation in the Wavelet domain (LIW) is designed, to reduce the low-level disturbance of the conditional encoder on diffusion noise estimation. The essence of infrared imaging cause that the infrared noise is global. The background radiation predominantly consists of low-frequency components, whereas the higher-intensity targets are evident in the high-frequency components \cite{li2020infrared}. This should be considered from two aspects.

First, the wavelet transform demonstrates favorable localization properties in both the time and frequency domains, as well as multi-scale features, facilitating image processing across various scales. The \textit{k}-level 2-D Haar Discrete Wavelet Transform (HDWT) \cite{haar} is applied to the original infrared features, as shown in Figure \ref{fig3} and Formula \ref{16}, to obtain the low-frequency approximation component $\mathbf{L}^k$ and the high-frequency vertical, horizontal, and diagonal components $\mathbf{H}^k_v, \mathbf{H}^k_h, \mathbf{H}^k_d$ $(k=1, 2, \dots K)$. This k-level 2-D HDWT of partial features is shown in Fig. \ref{fig4} and Fig. \ref{fig5}. A substantial presence of background noise interference is evident in the low-frequency component. Consequently, the low-frequency approximate component $\mathbf{L}^k$ $(k=1, 2, \dots K)$ is utilized for residual mapping, as shown in Formula \ref{18} and \ref{19}. Then, 2-D Haar Inverse Discrete Wavelet Transform (HIDWT) based on original high-frequency components $\mathbf{H}^k_v, \mathbf{H}^k_h, \mathbf{H}^k_d$ $(k=1, 2, … K)$ is applied to restore the infrared small target information, and initiated with the \textit{K}-level low-frequency approximate component $\mathbf{\tilde{L}}^K=\mathbf{f}^K_i$, as shown in Formula \ref{18}.

Second, Swin Block \cite{swin} is employed to process the global disturbances of infrared features. Transformers \cite{vaswani2017attention} are marvelous in capturing the semantic structure in images and have strong global perception. It is applied for global semantic compression in the wavelet domain, to isolate low-frequency disturbances while preserving the semantic structure of infrared features.
\begin{equation}
	\label{16}
	\mathbf{L}^k,{\mathbf{H}^k_v, \mathbf{H}^k_h, \mathbf{H}^k_d}=HDWT(\mathbf{f}^{k-1}_i),
\end{equation}

\begin{equation}
	\begin{aligned}
		\mathbf{f}^{k}_i=S(\mathbf{L}^k),\\
		k=1,2,\dots,K-1,
	\end{aligned}
\end{equation}
where, $HDWT(.)$ represents the 2-D Haar Discrete Wavelet Transform, $\mathbf{f}^{k}_i$ is the \textit{i}-th stage features under \textit{k}-level transform, and $S(.)$ denotes the Swin Block.

To obtain the optimal features $\hat{\mathbf{f}}$, it is easier to optimize the residual mapping $\mathbf{f}=\hat{\mathbf{f}}+\mathbf{R}$ when the original mapping $\mathbf{f}\to\hat{\mathbf{f}}$ is an identity-like mapping \cite{he2016deep, zhang2017beyond}. Finally, as shown in Figure \ref{fig3}, LIW estimates a residual $R$ to isolate low-frequency disturbances:
\begin{equation}
	\label{18}
	\begin{aligned}
		\mathbf{\tilde{L}}^{k-1}=Conv(HIDWT(\mathbf{\tilde{L}}^{k},{\mathbf{H}^k_v, \mathbf{H}^k_h, \mathbf{H}^k_d})),\\
		k=K,K-1,\dots,2,
	\end{aligned}
\end{equation}
\begin{equation}
	\label{19}
	\mathbf{R}=RO(\mathbf{\tilde{L}}^1),
\end{equation}
where, $HIDWT(.)$ represents the 2-D Haar Inverse Discrete Wavelet Transform, $Conv(.)$ is a convolutional module, $RO(.)$ represents the residual output block, and $\mathbf{\tilde{L}}^k$ is \textit{k}-level inversed low-frequency approximate component.

\begin{figure*}[t]
	\centering
	\includegraphics[width=0.75\textwidth]{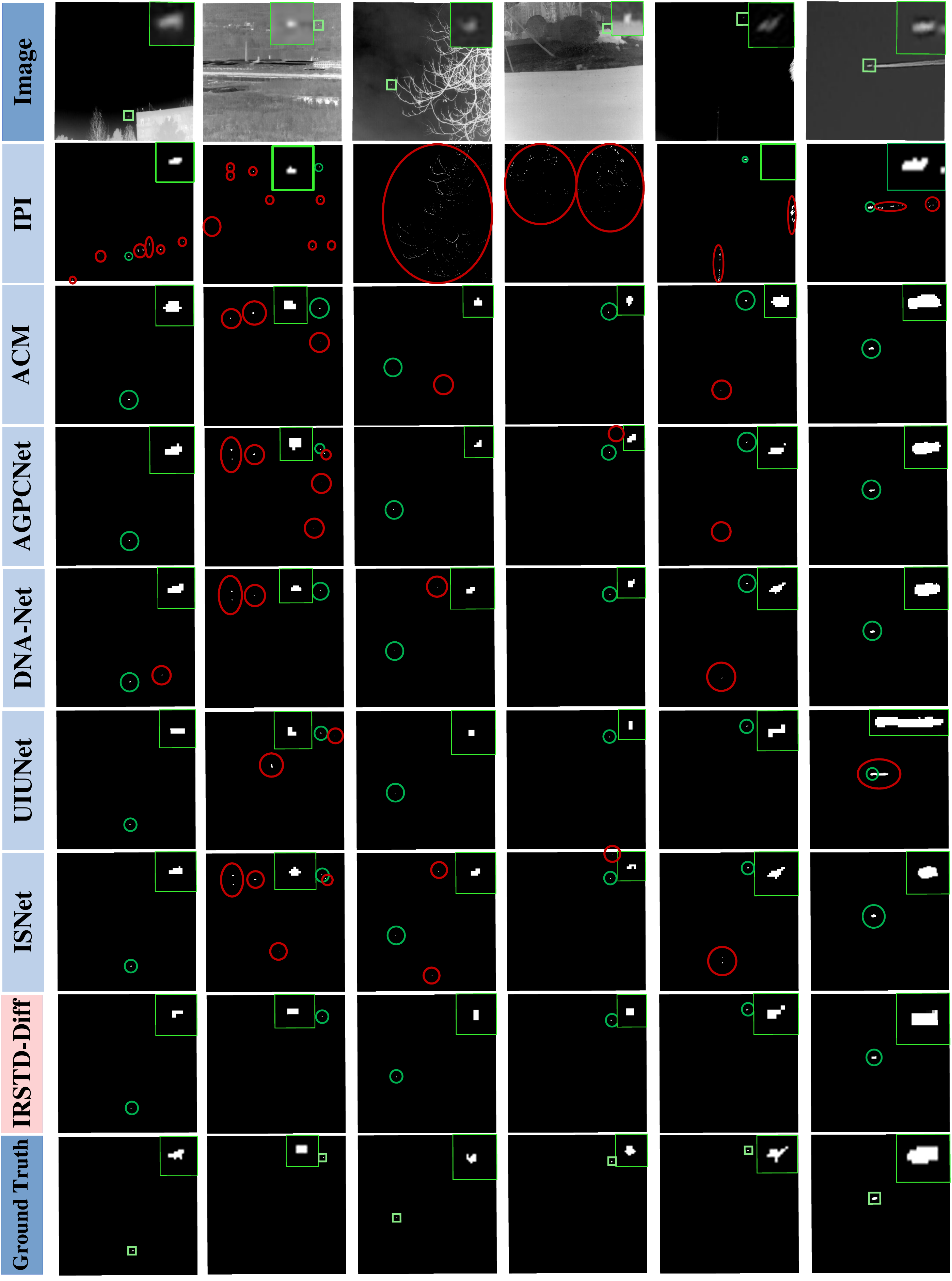}
	\caption{\textbf{Qualitative results obtained by different discriminative methods on NUAA-SIRST and IRSTD-1k datasets.} Enlarged targets are shown in the right-top corner. Circles in green, red, and yellow represent correctly detected targets, false alarm, and miss detected targets, respectively.}
	\label{fig9}
\end{figure*}

\begin{figure*}[t]
	\centering
	\includegraphics[width=0.9\textwidth]{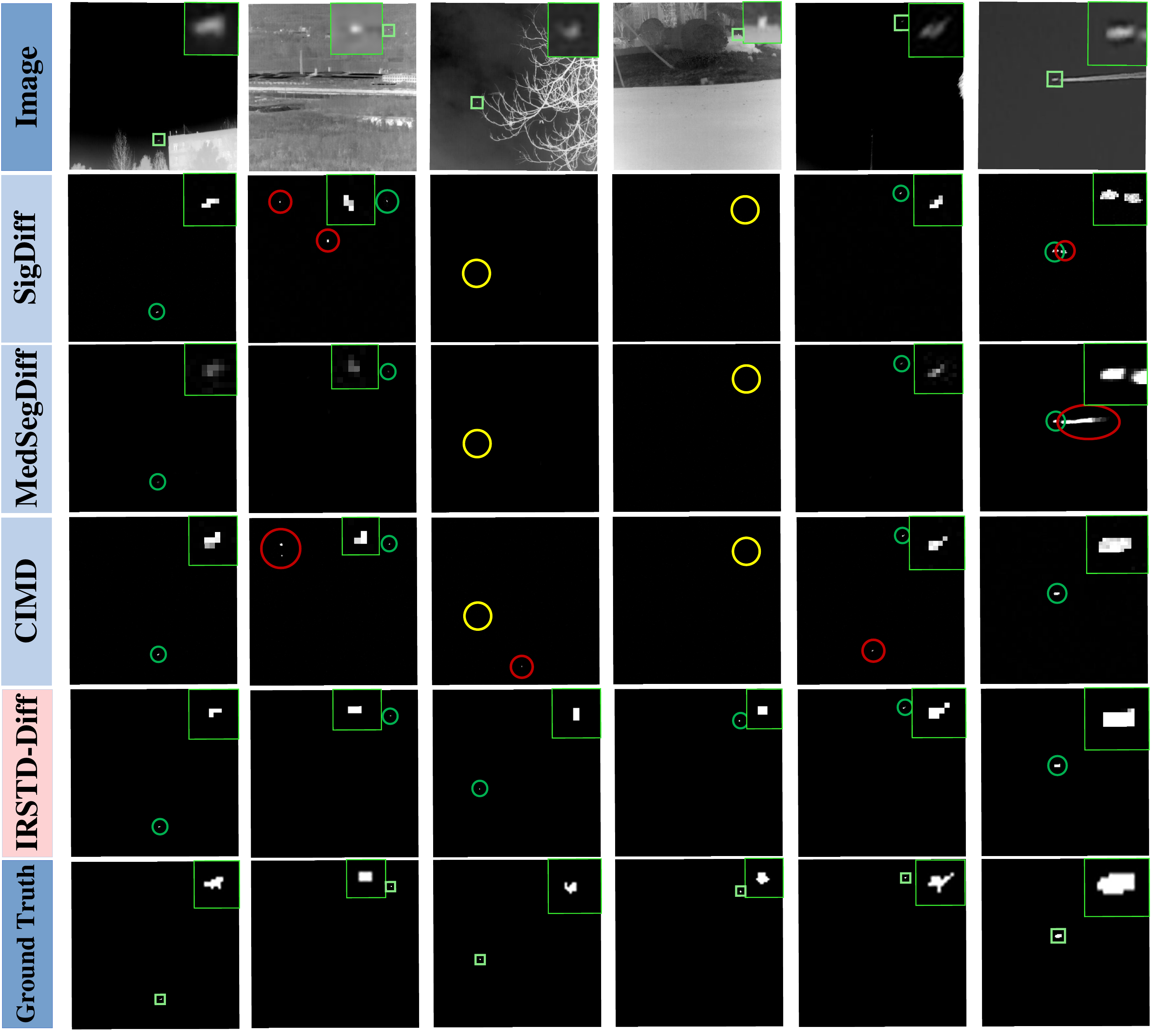}
	\caption{\textbf{Qualitative results obtained by diffusion-based generative methods on NUAA-SIRST and IRSTD-1k datasets.} Enlarged targets are shown in the right-top corner. Circles in green, red, and yellow represent correctly detected targets, false alarm, and miss detected targets, respectively.}
	\label{fig10}
\end{figure*}

\begin{table*}[!t]
	\centering
	\small
	\caption{\textbf{Unified configuration of the diffusion models.}}
	\label{t1}
	\begin{tabular}{ccccccccccc}
		\Xhline{1.2pt}
		\makecell[c]{Image\\Size}&\makecell[c]{Batch\\Size}&\makecell[c]{Diffusion\\Step}&\makecell[c]{Training\\Step}&\makecell[c]{UNet\\Channels}&Resblocks&Heads&\makecell[c]{Attention\\Resolutions}&\makecell[c]{Channel\\Multiple}&\makecell[c]{Input\\Channels}&\makecell[c]{Output\\Channels}\\ 
		\Xhline{1pt}
		256&4&100&80000&64&2&4&‘16’&(1, 1, 2, 2, 4, 4)&3&1\\
		\Xhline{1.2pt}
	\end{tabular} 
\end{table*}

\begin{table}[t]
	\centering
	\caption{\textbf{Detailed configuration of the Conditional Encoder (CE).} \textit{In\_C}, \textit{Mid\_C}, and \textit{Out\_C} represent the input, middle, and output channels, respectively. \textit{Up-Down} indicates whether to use up / down-sampling.}
	\label{t2}
	\renewcommand\arraystretch{1.2}
	\begin{tabular}{ccccc}
		\Xhline{1.2pt}
		Modules&\textit{In\_C}&\textit{Mid\_C}&\textit{Out\_C}&\textit{Up-Down}\\ 
		\Xhline{1pt}
		Encoder1&3&16&64&True\\
		Encoder2&64&16&64&True\\
		Encoder3&64&32&128&True\\
		Encoder4&128&32&256&True\\
		Encoder5&256&32&256&False\\
		Encoder6&256&64&256&False\\
		\Xhline{1pt}
		Decoder5&512&64&256&False\\
		Decoder4&512&32&128&True\\
		Decoder3&256&32&64&True\\
		Decoder2&128&16&64&True\\
		Decoder1&128&16&64&True\\
		\Xhline{1.2pt}
	\end{tabular} 
\end{table}

\begin{table*}[thbp]
	\centering
	\caption{\textbf{Comparisons with SOTA methods on NUAA-SIRST, IRSTD-1k, and NUDT-SIRST datasets in $IoU(\%)$, $Pd(\%)$ and $Fa(10^{-6})$}. The best results are in \textbf{BOLD}.}
	\label{t3}
	\renewcommand\arraystretch{1.5}
	\begin{tabular}{c|c|ccc|ccc|ccc}
		\Xhline{1.2pt}
		Methods&Description&\multicolumn{3}{c|}{NUAA-SIRST}&\multicolumn{3}{c|}{IRSTD-1k}&\multicolumn{3}{c}{NUDT-SIRST}\\
		&&$IoU$&$F_a\downarrow$&$P_d$&$IoU$&$F_a\downarrow$&$P_d$&$IoU$&$F_a\downarrow$&$P_d$\\ 
		\Xhline{1pt}
		WSLCM \cite{WSLCM}&Local Contrast Based&1.16&5446&77.95&3.45&6619&72.44&2.28&1309&56.82\\
		TLLCM \cite{TLLCM}&Local Contrast Based&11.03&7.27&79.47&5.36&4.93&63.97&7.06&46.12&62.01\\
		MSLSTIPI \cite{MSLSTIPT}&Local Rank Based&10.30&1131&82.13&10.37&3707&57.05&8.34&888.1&47.40\\
		PSTNN \cite{PSTNN}&Local Rank Based&22.40&29.11&77.95&24.57&35.26&71.99&14.85&44.17&66.13\\
		IPI \cite{IPI}&Local Rank Based&25.67&11.47&85.55&27.92&16.18&81.37&17.76&41.23&74.49\\
		\Xhline{1pt}
		ACM \cite{ACM}&Discriminative CNN Based&63.06&4.04&94.50&55.38&22.21&83.22&65.94&5.60&94.09\\
		AGPC \cite{AGPC}&Discriminative CNN Based&69.08&4.48&98.17&61.00	&12.96&88.36&84.49&2.69&97.58\\
		ISNet \cite{ISNet}&Discriminative CNN Based&71.12&7.14&98.17&64.47&11.52&92.12&81.35&4.66&96.77\\
		DNANet \cite{DNA}&Discriminative CNN Based&69.85&8.38&98.17&64.44&8.98&92.12&89.57&2.72&98.12\\
		UIUNet \cite{uiunet}&Discriminative CNN Based&\textbf{75.30}&8.83&\textbf{99.08}&63.68&10.49&92.81&91.98&2.92&\textbf{98.66}\\
		\Xhline{1pt}
		SigDiff \cite{sigdiff}&Diffusion Based&66.00&4.17&94.50&47.41&8.67&81.16&75.53&4.72&94.89\\
		MedSegDiff \cite{medsegdiff2}&Diffusion Based&64.19&16.80&97.25&57.65&21.94&81.16&78.27&10.15&96.50\\
		CIMD \cite{CIMD}&Diffusion Based&71.00&3.68&97.25&62.32&6.84&87.33&89.15&1.48&97.31\\
		\rowcolor{gray!30}\textbf{IRSTD-Diff(ours)}&Diffusion Based&74.09&\textbf{1.46}&\textbf{99.08}&\textbf{65.71}&\textbf{5.80}&\textbf{93.15}&\textbf{92.14}&\textbf{1.08}&98.12\\
		\Xhline{1.2pt}
	\end{tabular} 
\end{table*}

\section{Experiments}

\subsection{Setup}
\subsubsection{Datasets}
Our experiments are conducted on NUAA-SIRST \cite{ACM}, IRSTD-1k \cite{ISNet} and NUDT-SIRST \cite{DNA} datasets. The NUAA-SIRST dataset has 341 training data and 86 testing data. The IRSTD-1k dataset has 800 training data and 201 testing data. And the NUDT-SIRST dataset has 1062 training data and 265 testing data. We use the same division setting as original paper \cite{ACM, ISNet, DNA}.

\subsubsection{Evaluation Metrics}
\textit{1) Intersection over Union (IoU):} IoU, a common pixel-level evaluation metric of IRSTD, is the ratio of the intersection and union region between the predictions and the ground truth. IoU is calculated on the whole dataset and the correctness of each pixel has a great impact.
\begin{equation}
	\label{12}
	IoU=\frac{A_i}{A_u}=\frac{\sum\limits_{i=1}^{n}{TP}_i}{\sum\limits_{i=1}^{n}T_i+P_i-{TP}_i},
\end{equation}
where $A_i$ and $A_u$ are the intersection and union, respectively. $T$ denotes the pixels predicted as the targets. $P$ denotes the pixels of the ground truth targets. $TP$ is the true positive pixels. $n$ represent the number of infrared images in the test set.

\textit{2) Probability of Detection ($P_d$)}: $P_d$ evaluates the ratio of the true targets detected by the model to all the ground truth targets. It is a target-level evaluation metric. A detected target is considered as $TP$ if the centroid derivation is less than 3 \cite{DNA}.
\begin{equation}
	\label{14}
	P_d=\frac{{TP}_{sum}}{T_{sum}},
\end{equation}
where ${TP}_{sum}$ represents the number of detected targets, $T_{sum}$ represents all the targets in the test set.

\textit{3) False-Alarm Rate (Fa)}: A target-level evaluation metric, to measure the ratio of predicted target pixels that do not match the ground truth.
\begin{equation}
	\label{15}
	F_a=\frac{\sum\limits_{i=1}^{n}{FP}_{i}}{ALL},
\end{equation}
where $FP$ denotes the number of false alarm pixels, and $ALL$ is all pixels in the image.

\subsubsection{Implementation Details}
AdamW \cite{loshchilov2018decoupled} is utilized as the optimizer. Weight decay is 0 and the learning rate is 0.0001, with a linear learning rate strategy. As the configuration of the diffusion model shown in Table \ref{t1}, an inferior configuration is used to verify the feasibility, limited by the computility. The implementation of a more generalized diffusion model configuration may yield additional enhancements in performance, but it necessitates elevated standards for computility. The data pattern of the infrared small target masks is oversimplified compared with standard visual images, with weak discreteness. Accordingly, the \textit{step} is set to 100. It can be reduced to an smaller value (e.g., 30) to diminish sampling time and enhance efficiency. The input images are resized to 256 × 256 during training and testing. The IRSTD-Diff is trained for 80000 steps.

\textit{Comparative discriminative methods:} For the factors outside the model, such as learning rate strategy, data augmentation, weight initialization, training strategy, etc., are unified to make a fair comparison \cite{ACM}. All the methods are trained 400 epochs with a batch size of 8.

\textit{Comparative generative methods:} Use the unified configuration as Table \ref{t1}. Select 8 different steps for testing and choose the best result.

All methods are implemented based on Pytorch 1.10.0 and conducted on an NVIDIA GeForce RTX 3080 Ti.

Detailed configuration of the Conditional Encoder (CE) is
shown as Table \ref{t2}. All experiments in this paper are completed with this configuration.

\subsection{Quantitative comparisons}
The IRSTD-Diff is compared with 13 SOTA methods including unlearnable methods: WSLCM \cite{WSLCM}, TLLCM \cite{TLLCM}, MSLSTIPT \cite{MSLSTIPT}, PSTNN \cite{PSTNN}, IPI \cite{IPI}; discriminative methods: ACM \cite{ACM}, AGPCNet \cite{AGPC}, DNANet \cite{DNA}, ISNet \cite{ISNet}, UIUNet\cite{uiunet}; and generative methods: SigDiff \cite{sigdiff}, MedSegDiff \cite{medsegdiff2}, CIMD \cite{CIMD}.

As shown in Table \ref{t2}, Non-learning methods exhibit significant disadvantages compared to deep learning-based approaches. On the NUDT-SIRST dataset, the pixel-level accuracy (IoU) shows a disparity of -74.38 \%, and the target-level accuracy has differences of 40.15 × $10^{-6}$ Fa and -24.17 \% Pd. At present, non-learning paradigms have none advantage compared to the data-driven deep learning paradigms.

Our IRSTD-Diff attains state-of-the-art performance on three datasets. As shown in Table \ref{t3}, our IRSTD-Diff is excellent in the target-level performance compared with other discriminative methods. It is evident that our IRSTD-Diff has established substantial superiority through the comparison of false alarm rate. In contrast to DNANet, which excels in pixel-level performance, our approach results in reductions of 7.37 × $10^{-6}$, 3.18 × $10^{-6}$, and 1.64 × $10^{-6}$ on the three datasets. In terms of  probability of detection, our IRSTD-Diff demonstrates comparable performance to UIUNet on the NUAA-SIRST and NUTD-SIRST datasets, and surpasses it on the IRSTD-1k dataset. Our method is comprehensively superior to other discriminative models.

The aforementioned observations serve to validate the assumptions articulated in the Section \ref{sec1}. At the later period of discriminative training, it is difficult to fluctuate the experience risk, owing to a small proportion of false alarm and missed detection pixels. It is even more demanding to punish this phenomenon because of the decayed learning rate. Our approach compensates for this issue from the perspective of generative learning, i.e., modeling the mask posterior distribution. The scattered false alarm pixels bring strong disturbances to the simple data pattern. Therefore, KL divergence constraint on the mask posterior distribution can suppress such outliers. Therefore, IRSTD-Diff has low false alarm and missed detection, and competitive pixel-level performance. Five of the experimental outcomes outperformed the state-of-the-art methods in the six comparison experiments conducted on the three datasets. Besides, IRSTD-Diff has competitive pixel-level performance. With the exception of a slightly 1.21 \% lower IoU on the NUAA-SIRST dataset compared to UIUNet, the IRSTD-Diff achieves the state-of-the-art performance on the other two datasets, with improvements of 1.24 \% IoU compared to ISNet and 0.16 \% IoU to UIUNet.

Similarly, our IRSTD-Diff has achieved comprehensive advantages compared with other diffusion-based methods. The methods we compared with is the latest diffusion-based semantic segmentation or medical image segmentation approaches \cite{sigdiff, medsegdiff2, CIMD}. These excellent diffusion-based methods have made a pioneering contribution to the discriminative DDPM application. However, it is difficult to directly apply to IRSTD owing to the absence of pertinence design, resulting in inferior pixel-level and Pd performance. However, we observe a counter-intuitive anomaly: these non-pertinence designed, diffusion-based dense prediction methods attain competitive false alarm rate. The false alarm rate of CIMD \cite{CIMD} surpassed state-of-the-art discriminative methods on three datasets, and SigDiff \cite{sigdiff} also delivered performance approaching the state-of-the-art methods. This atypical phenomenon further corroborates the hypothesis proposed in the Section \ref{sec1}, i.e., generative diffusion-based approaches effectively restrain minimal false alarm pixels that would not significantly disturb empirical risk.

In this work, the diffusion model for IRSTD is pertinence improved, which makes the diffusion model truly applied to the issue of IRSTD. Our IRSTD-Diff demonstrates superior performance at the target-level, particularly in Pd, with improvements of 1.83 \%, 5.82 \%, and 0.81 \% compared to the optimal CIMD. Our pertinence enhancements specifically designed to the intrinsic characteristics of infrared small targets, significantly heighten pixel-level accuracy, yielding 3.09 \%, 3.39 \%, and 2.99 \% IoU increasements on the three datasets, respectively. These results establish the effectiveness of our generative diffusion-based approach and emphasize the applicability for infrared small target detection. 

The ROC curves shown in Fig. \ref{fig8} demonstrate the superiority and stability of our IRSTD-Diff. We quantitatively compare our IRSTD-Diff with other SOTA methods in AUC, as shown in Table \ref{t4}. IRSTD-Diff has better performance than unlearnable methods and nonpertinent designed generative methods. It also has competitiveness compared with the discriminant methods.

\begin{table*}[t]
	\centering
	\caption{\textbf{Area Under Curve (AUC) comparision with other methods.}}
	\label{t4}
	\renewcommand\arraystretch{1.2}
	\begin{tabular}{ccccccccccc}
		\Xhline{1.2pt}
		Methods&IPI&ACM&AGPC&DNA&ISNet&UIUNet&SigDiff&MedSegDiff&CIMD&IRSTD-Diff\\
		\Xhline{1pt}
		AUC&0.4545&0.6924&0.7019&0.7534&0.7692&0.8303&0.6727&0.6680&0.7828&0.7962\\
		\Xhline{1.2pt}
	\end{tabular} 
\end{table*}

\subsection{Qualitative comparisons}
Qualitative results are obtained by different methods on NUAA-SIRST and IRSTD-1k datasets. Most methods prefer to find relatively bright spots in the infrared images, owing to the dataset propensity. This brings numerous false alarm. Our IRSTD-Diff makes a better penalty for such a tendency, as shown in Fig. \ref{fig9} and Fig. \ref{fig10}, thus achieving a significantly low false alarm. Meanwhile, its performance in missed detection and pixel-level edge fitting is competitive. UIUNet demonstrates a relatively coarse fit to the edges, presenting targets in block-like pixel clusters. In contrast, DNANet and ISNet focus on target edges without block-like clusters, but suffer from excessive smoothness which impacts pixel-level performance. Our approach achieves a better compromise.

The proclivity of results from diffusion-based methods exhibits fundamental similarity, as shown in Fig. \ref{fig10}. These methods evince significantly fewer false alarms compared to the discriminative models in Fig. \ref{fig9}. Thereby the effectiveness of our approach in addressing the hypothesis problem (Section \ref{sec1}) is confirmed. However, there is a massive amount of miss detection, with lower Pd than state-of-the-art discriminative methods by more than 10 \%. Naturally, this results in substantial pixel-level deficiencies. The subpar pixel-level performance is not due to low accuracy in individual samples. On the contrary, the detected samples are precise in pixel-level. Given the prevalence of miss detection, resulting in 0 IoU for a considerable number of samples, the pixel-level performance is inevitably impacted. Our approach effectively averts this predicament, thereby facilitating the feasibility of the diffusion-based paradigm in IRSTD.

The low-frequency isolation module in the wavelet domain (LIW), as shown in Fig. \ref{fig4}, conspicuously isolates the background radiation and obtain a cleaner infrared conditional encoder. It makes the conditional encoder network attach significance to the target area, which greatly reduces its attention dispersivity compared with no LIW module.

\section{Analysis}
\label{sec5}
\subsection{How do the \textit{k} of LIW affects the model’s performance?}
As shown in Table \ref{t3}, we discuss the value of $\mathit{k}$ in the $\mathit{k}$-level 2-D HDWT. Wavelet transform, a double reduced resolution transform, makes the LIW enormously susceptible to \textit{k}. Excessive wavelet transform with $\mathit{k}>3$ will still lose several information because of the small target size, resulting in inferior performance. When $\mathit{k}=1$, the module capacity is not enough to fit a better residual $\mathbf{R}$. In summary, the best $\mathit{k}$ (resolution dependent) is 2 for 256 resolution.

\subsection{How do the diffusion training steps affect the model’s performance?}
The mask data pattern is uncomplicated, thereupon the diffusion training steps can be smaller than the original configuration (1000) \cite{DDPM}. As shown in Table \ref{t4}, taking 100 training steps has achieved a stupendous result; Under excessive steps, the evaluation metrics fluctuate, but there is no obvious improvement; And too few steps are not feasible.

\begin{figure}[t]
	\centering
	\includegraphics[width=0.5\textwidth]{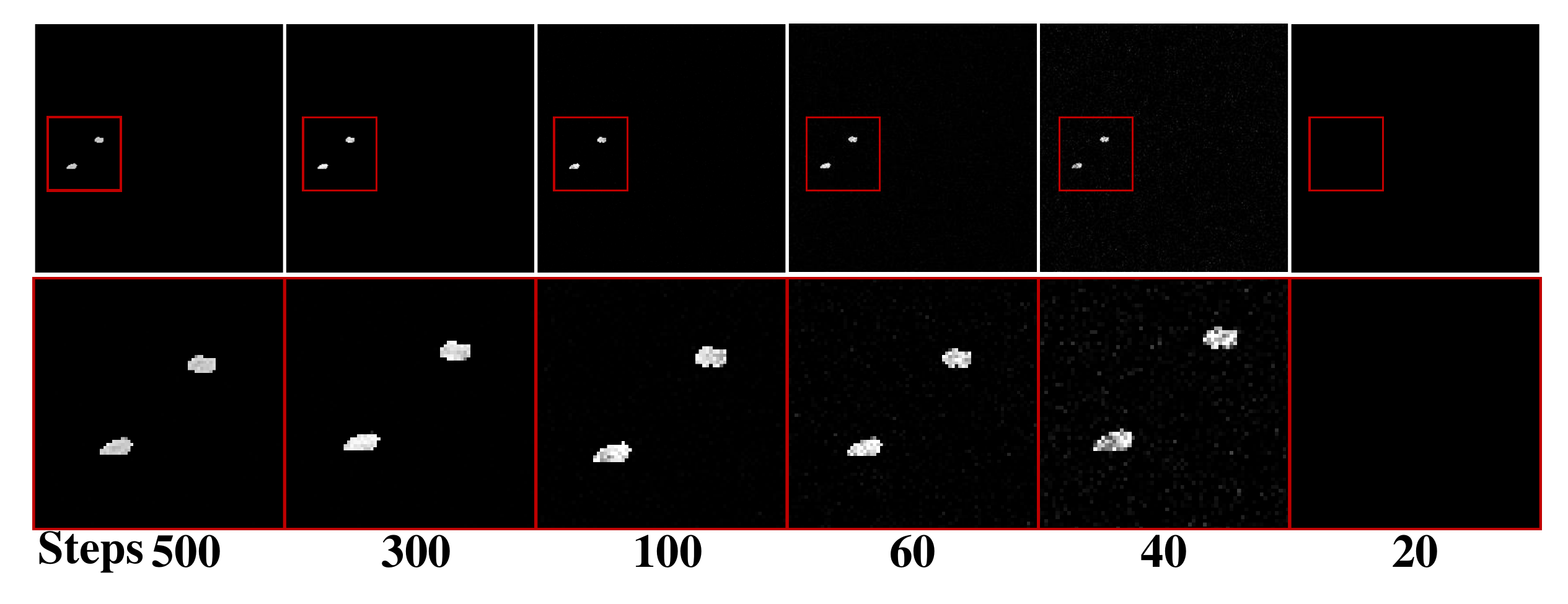}
	\caption{\textbf{Undersampling and oversampling with various steps.} The second line is the magnification of the red box area in the first line.}
	\label{fig11}
\end{figure}

\begin{figure}[t]
	\centering
	\includegraphics[width=0.5\textwidth]{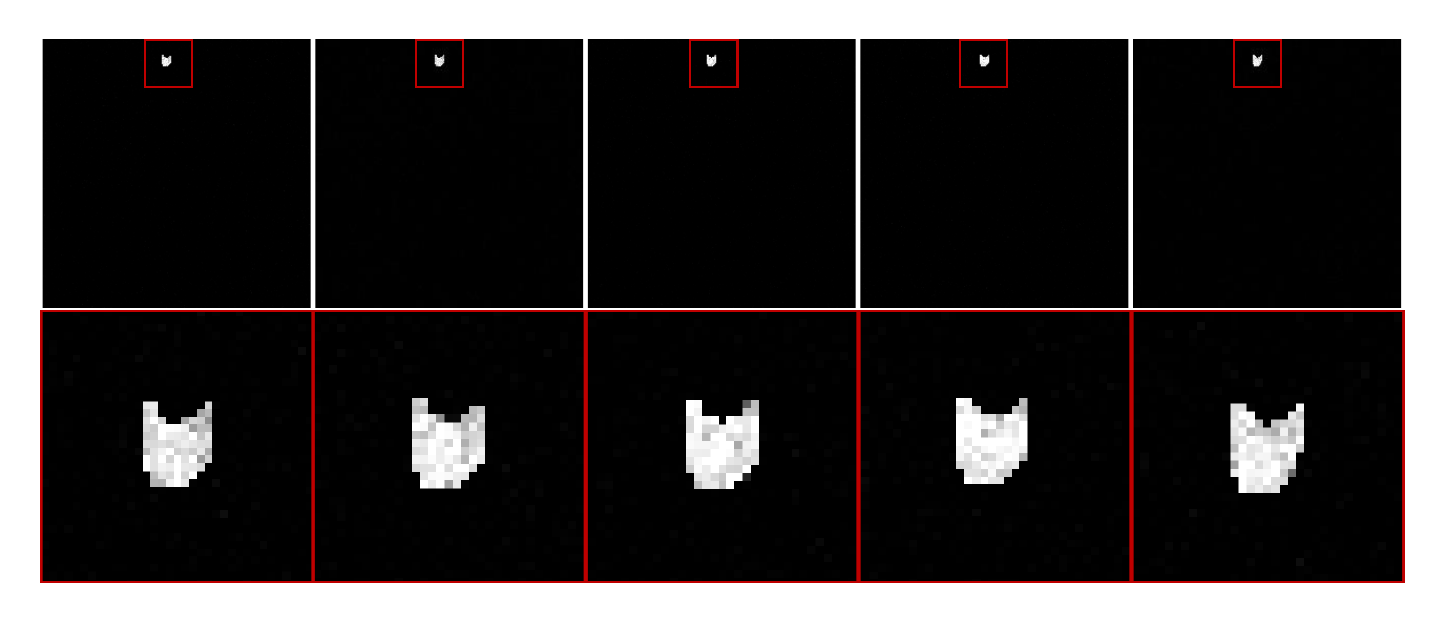}
	\caption{\textbf{Multiple samples under an identical posterior distribution.} The second line is the magnification of the red box area in the first line. The results demonstrate the stability of sampling.}
	\label{fig12}
\end{figure}

\subsection{Undersampling or Oversampling}
The $\beta$ is large at early steps, and the reverse process is intense, while the latter steps are converse \cite{DDPM}. This tendency is reflected in Figure \ref{fig6}. To foreshorten the inference sampling time, undersampling can be adopted. As shown in Table \ref{t5} and Figure \ref{fig6}, 60 steps can achieve a reasonable performance. Oversampling does not bring any improvement.

\subsection{What are the contributions of CE and LIW, respectively?}
We performed ablation studies on the IRSTD-1k dataset, to verify the effectiveness of the latent space encoder $\boldsymbol{\mathcal{I}}'$, Low-frequency Isolation in the Wavelet domain and baseline, as shown in Table \ref{t7}. All the above modules can improve the performance of the baseline.

\subsection{Sample stability}
The infrared small targets are small in size and simple in appearance, and consequently, the differential of multiple sampling results is not significant, as shown in Figure \ref{fig7}. As quantitatively shown in Table \ref{t6}, the Ensemble \cite{warfield2004simultaneous} is not necessary for the stability samples, with considering the sampling time.

\begin{table}[!t]
	\centering
	\caption{\textbf{Effects of the $k$ in LIW.} The training step is set to 100, and we stochastically choose a stationary posterior probability (60000 steps) for inference.}
	\label{t5}
	\renewcommand\arraystretch{1.5}
	\begin{tabular}{ccccc}
		\Xhline{1.2pt}
		Steps&4&3&2&1\\ 
		\Xhline{1pt}
		$IoU(\%)$&62.06&62.17&64.98&61.80\\
		$F_a(10^{-6})\downarrow$&8.17&6.37&7.42&6.00\\
		$P_d(\%)$&91.10&91.10&93.15&91.44\\
		\Xhline{1.2pt}
	\end{tabular} 
\end{table}

\begin{table}[!t]
	\centering
	\caption{\textbf{Effects of the diffusion training steps.} Small steps can achieve stupendous performance. Limited by $\beta$, step cannot be less than 20. \textit{NW} means Not Working.}
	\label{t6}
	\renewcommand\arraystretch{1.5}
	\begin{tabular}{cccccccc}
		\Xhline{1.2pt}
		Steps&1000&500&200&100&50&30&20\\ 
		\Xhline{1pt}
		$IoU$&64.41&63.32&64.28&65.71&61.32&59.43&\textit{NW}\\
		$F_a\downarrow$&8.58&8.13&7.96&5.80&2.59&5.61&\textit{NW}\\
		$P_d$&93.84&92.12&93.15&93.15&89.38&91.10&\textit{NW}\\
		\Xhline{1.2pt}
	\end{tabular} 
\end{table}

\begin{figure*}[t]
	\centering
	\includegraphics[width=1\textwidth]{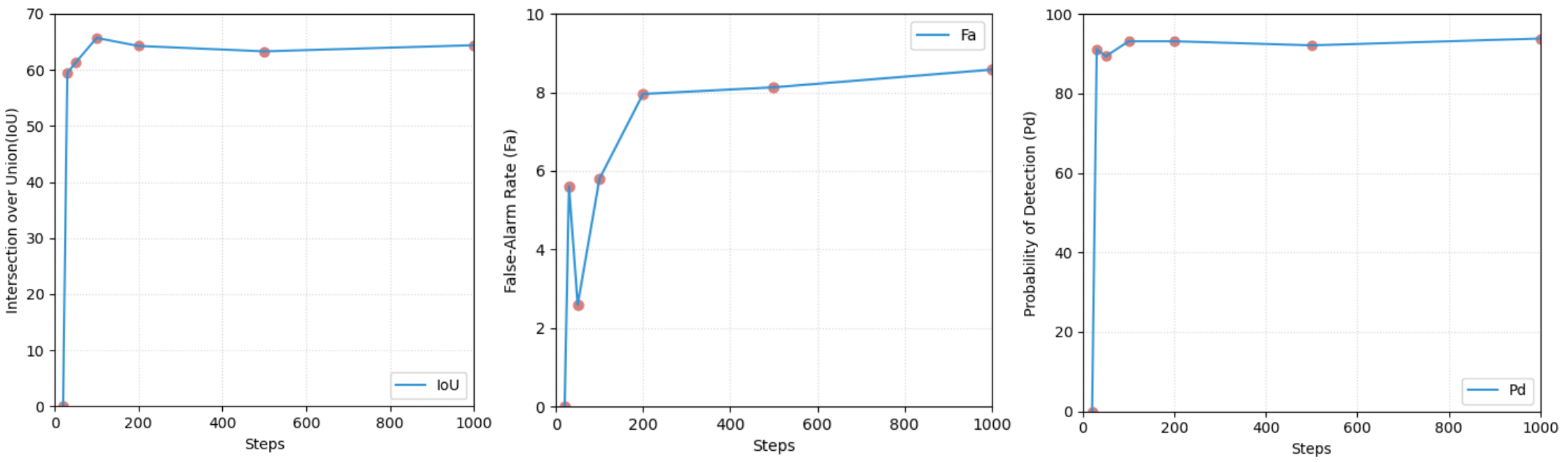}
	\caption{\textbf{Effects of the diffusion training steps.} Small steps can achieve stupendous performance. When the step is greater than 100, continuing to increase the step is not beneficial to performance improvement, but oscillates in a small range.}
	\label{line}
\end{figure*}

\begin{table}[!t]
	\centering
	\caption{\textbf{Undersampling and oversampling during inference.} The training step is set to 100, and we stochastically choose a stationary posterior probability (60000 steps) for inference.}
	\label{t7}
	\renewcommand\arraystretch{1.5}
	\begin{tabular}{ccccccc}
		\Xhline{1.2pt}
		Steps&500&300&100&80&60&40\\ 
		\Xhline{1pt}
		$IoU(\%)$&64.93&64.76&64.98&64.61&64.42&31.66\\
		$Fa(10^{-6})\downarrow$&8.74&8.24&7.42&7.06&7.00&51.92\\
		$Pd(\%)$&93.15&93.15&93.15&93.15&93.15&92.81\\
		\Xhline{1.2pt}
	\end{tabular} 
\end{table}

\begin{table}[!t]
	\centering
	\caption{\textbf{Effects of the Ensemble.}}
	\label{t8}
	\renewcommand\arraystretch{1.5}
	\begin{tabular}{cccc}
		\Xhline{1.2pt}
		Steps&$IoU(\%)$&$P_d(\%)$&$F_a(10^{-6})\downarrow$\\ 
		\Xhline{1pt}
		5&64.86&93.15&7.19\\
		3&64.55&93.15&7.31\\
		1&64.98&93.15&7.42\\
		\Xhline{1.2pt}
	\end{tabular} 
\end{table}

\begin{table}[!t]
	\centering
	\caption{\textbf{Ablation: Contribution of our proposed CE and LIW.}}
	\label{t9}
	\renewcommand\arraystretch{1.5}
	\begin{tabular}{ccccccc}
		\Xhline{1.2pt}
		\#&NE&CE&LIW&$IoU(\%)$&$F_a(10^{-6})\downarrow$&$P_d(\%)$\\ 
		\Xhline{1pt}
		0& &\checkmark& &65.07&7.63&88.70\\
		1& &\checkmark&\checkmark&60.30&8.74&90.07\\
		2&\checkmark& & &58.85&12.57&86.64\\
		3&\checkmark&\checkmark& &64.27&5.68&91.10\\
		4&\checkmark&\checkmark&\checkmark&65.69&5.80&93.15\\
		\Xhline{1.2pt}
	\end{tabular} 
\end{table}

\section{Conclusion}
In this paper, we have presented a diffusion model framework for Infrared Small Target Detection, IRSTD-Diff, from the perspective of mask posterior distribution generating. The target-level insensitivity of the vanilla discriminative model is ameliorated. In addition, a low-frequency isolation module in the wavelet domain is designed, to reduce the impact of low-level interference infrared features on diffusion noise estimation.
IRSTD-Diff is an efficacious generative approach for IRSTD. Experimental results on three datasets demonstrate that IRSTD-Diff achieves superior performance against other diffusion-based and discriminative methods. However, diffusion-based methods have inferior inference sampling speed than discriminative methods, which entails additional sampling efficiency improvement in the future. Besides, the diffusion-based methods are difficult to save the optimal results during training, which limits our approach.

\bibliographystyle{IEEEtran}
\bibliography{TGRS}

\begin{thebibliography}{10}
\providecommand{\url}[1]{#1}
\csname url@samestyle\endcsname
\providecommand{\newblock}{\relax}
\providecommand{\bibinfo}[2]{#2}
\providecommand{\BIBentrySTDinterwordspacing}{\spaceskip=0pt\relax}
\providecommand{\BIBentryALTinterwordstretchfactor}{4}
\providecommand{\BIBentryALTinterwordspacing}{\spaceskip=\fontdimen2\font plus
\BIBentryALTinterwordstretchfactor\fontdimen3\font minus
  \fontdimen4\font\relax}
\providecommand{\BIBforeignlanguage}[2]{{%
\expandafter\ifx\csname l@#1\endcsname\relax
\typeout{** WARNING: IEEEtran.bst: No hyphenation pattern has been}%
\typeout{** loaded for the language `#1'. Using the pattern for}%
\typeout{** the default language instead.}%
\else
\language=\csname l@#1\endcsname
\fi
#2}}
\providecommand{\BIBdecl}{\relax}
\BIBdecl

\bibitem{lu2006detecting}
J.~Lu, Y.~He, H.~Li, and F.~Lu, ``Detecting small target of ship at sea by
  infrared image,'' in \emph{2006 IEEE International Conference on Automation
  Science and Engineering}, 2006, pp. 165--169.

\bibitem{zhao2022single}
M.~Zhao, W.~Li, L.~Li, J.~Hu, P.~Ma, and R.~Tao, ``Single-frame infrared
  small-target detection: A survey,'' \emph{IEEE Geoscience and Remote Sensing
  Magazine}, vol.~10, no.~2, pp. 87--119, 2022.

\bibitem{demosthenous2015infrared}
P.~Demosthenous, C.~Pitris, and J.~Georgiou, ``Infrared fluorescence-based
  cancer screening capsule for the small intestine,'' \emph{IEEE transactions
  on biomedical circuits and systems}, vol.~10, no.~2, pp. 467--476, 2015.

\bibitem{deshpande1999max}
S.~D. Deshpande, M.~H. Er, R.~Venkateswarlu, and P.~Chan, ``Max-mean and
  max-median filters for detection of small targets,'' in \emph{Signal and Data
  Processing of Small Targets 1999}, vol. 3809, 1999, pp. 74--83.

\bibitem{Tophat}
J.-F. Rivest and R.~Fortin, ``Detection of dim targets in digital infrared
  imagery by morphological image processing,'' \emph{Optical Engineering},
  vol.~35, no.~7, pp. 1886--1893, 1996.

\bibitem{TLLCM}
J.~Han, S.~Moradi, I.~Faramarzi, C.~Liu, H.~Zhang, and Q.~Zhao, ``A local
  contrast method for infrared small-target detection utilizing a tri-layer
  window,'' \emph{IEEE Geoscience and Remote Sensing Letters}, vol.~17, no.~10,
  pp. 1822--1826, 2019.

\bibitem{WSLCM}
J.~Han, S.~Moradi, I.~Faramarzi, H.~Zhang, Q.~Zhao, X.~Zhang, and N.~Li,
  ``Infrared small target detection based on the weighted strengthened local
  contrast measure,'' \emph{IEEE Geoscience and Remote Sensing Letters},
  vol.~18, no.~9, pp. 1670--1674, 2020.

\bibitem{zhang2019infrared}
L.~Zhang and Z.~Peng, ``Infrared small target detection based on partial sum of
  the tensor nuclear norm,'' \emph{Remote Sensing}, vol.~11, no.~4, p. 382,
  2019.

\bibitem{sun2020infrared}
Y.~Sun, J.~Yang, and W.~An, ``Infrared dim and small target detection via
  multiple subspace learning and spatial-temporal patch-tensor model,''
  \emph{IEEE Transactions on Geoscience and Remote Sensing}, vol.~59, no.~5,
  pp. 3737--3752, 2020.

\bibitem{ALC}
Y.~Dai, Y.~Wu, F.~Zhou, and K.~Barnard, ``Attentional local contrast networks
  for infrared small target detection,'' \emph{IEEE Transactions on Geoscience
  and Remote Sensing}, vol.~59, no.~11, pp. 9813--9824, 2021.

\bibitem{uiunet}
X.~Wu, D.~Hong, and J.~Chanussot, ``Uiu-net: U-net in u-net for infrared small
  object detection,'' \emph{IEEE Transactions on Image Processing}, vol.~32,
  pp. 364--376, 2022.

\bibitem{ACM}
Y.~Dai, Y.~Wu, F.~Zhou, and K.~Barnard, ``Asymmetric contextual modulation for
  infrared small target detection,'' in \emph{IEEE/CVF Winter Conference on
  Applications of Computer Vision}, 2021, pp. 950--959.

\bibitem{DNA}
B.~Li, C.~Xiao, L.~Wang, Y.~Wang, Z.~Lin, M.~Li, W.~An, and Y.~Guo, ``Dense
  nested attention network for infrared small target detection,'' \emph{IEEE
  Transactions on Image Processing}, vol.~32, pp. 1745--1758, 2022.

\bibitem{ISNet}
M.~Zhang, R.~Zhang, Y.~Yang, H.~Bai, J.~Zhang, and J.~Guo, ``Isnet: Shape
  matters for infrared small target detection,'' in \emph{IEEE/CVF Conference
  on Computer Vision and Pattern Recognition}, 2022, pp. 877--886.

\bibitem{yu2016unitbox}
J.~Yu, Y.~Jiang, Z.~Wang, Z.~Cao, and T.~Huang, ``Unitbox: An advanced object
  detection network,'' in \emph{ACM International Conference on Multimedia},
  2016, pp. 516--520.

\bibitem{ILNet}
H.~Li, J.~Yang, Y.~Xv, and R.~Wang, ``Ilnet: Low-level matters for salient
  infrared small target detection,'' \emph{ArXiv}, p. abs/2309.13646, 2023.

\bibitem{vollmer2021infrared}
M.~Vollmer, ``Infrared thermal imaging,'' in \emph{Computer Vision: A Reference
  Guide}, 2021, pp. 666--670.

\bibitem{2005Noise}
H.~Jian-Jiang, L.~Zhao-Hui, and L.~Wen, ``Noise analysis of infrared image and
  muti-dim-small target's enhancement,'' \emph{Infrared Technology}, 2005.

\bibitem{malfait1997wavelet}
M.~Malfait and D.~Roose, ``Wavelet-based image denoising using a markov random
  field a priori model,'' \emph{IEEE Transactions on image processing}, vol.~6,
  no.~4, pp. 549--565, 1997.

\bibitem{jansen1999multiple}
M.~Jansen and A.~Bultheel, ``Multiple wavelet threshold estimation by
  generalized cross validation for images with correlated noise,'' \emph{IEEE
  transactions on image processing}, vol.~8, no.~7, pp. 947--953, 1999.

\bibitem{strela1999application}
V.~Strela, P.~N. Heller, G.~Strang, P.~Topiwala, and C.~Heil, ``The application
  of multiwavelet filterbanks to image processing,'' \emph{IEEE Transactions on
  image processing}, vol.~8, no.~4, pp. 548--563, 1999.

\bibitem{weyrich1998wavelet}
N.~Weyrich and G.~T. Warhola, ``Wavelet shrinkage and generalized cross
  validation for image denoising,'' \emph{IEEE Transactions on Image
  Processing}, vol.~7, no.~1, pp. 82--90, 1998.

\bibitem{haar}
A.~Haar, ``Zur theorie der orthogonalen funktionensysteme,''
  \emph{Mathematische Annalen}, vol.~69, no.~3, pp. 331--371, 1910.

\bibitem{li2020infrared}
J.~Li, P.~Zhang, X.~Wang, and S.~Huang, ``Infrared small-target detection
  algorithms: a survey,'' \emph{Journal of Image and Graphics}, vol.~25, no.~9,
  pp. 1739--1753, 2020.

\bibitem{AGPC}
T.~Zhang, S.~Cao, T.~Pu, and Z.~Peng, ``Agpcnet: Attention-guided pyramid
  context networks for infrared small target detection,'' \emph{IEEE Trans.
  Aerosp. Electron. Syst.}, vol.~59, no.~4, pp. 4256--4261, 2023.

\bibitem{DDPM}
J.~Ho, A.~Jain, and P.~Abbeel, ``Denoising diffusion probabilistic models,''
  \emph{Advances in neural information processing systems}, vol.~33, pp.
  6840--6851, 2020.

\bibitem{GAN1}
I.~Goodfellow, J.~Pouget-Abadie, M.~Mirza, B.~Xu, D.~Warde-Farley, S.~Ozair,
  A.~Courville, and Y.~Bengio, ``Generative adversarial networks,''
  \emph{Communications of the ACM}, vol.~63, no.~11, pp. 139--144, 2020.

\bibitem{GAN2}
T.~Karras, S.~Laine, and T.~Aila, ``A style-based generator architecture for
  generative adversarial networks,'' in \emph{IEEE/CVF Conference on Computer
  Vision and Pattern Recognition}, 2019, pp. 4401--4410.

\bibitem{VAE1}
D.~P. Kingma and M.~Welling, ``Auto-encoding variational bayes,'' \emph{ArXiv},
  p. abs/1312.6114, 2013.

\bibitem{VAE2}
H.~Kim and A.~Mnih, ``Disentangling by factorising,'' in \emph{International
  Conference on Machine Learning}, 2018, pp. 2649--2658.

\bibitem{liu2023more}
X.~Liu, D.~H. Park, S.~Azadi, G.~Zhang, A.~Chopikyan, Y.~Hu, H.~Shi,
  A.~Rohrbach, and T.~Darrell, ``More control for free! image synthesis with
  semantic diffusion guidance,'' in \emph{IEEE/CVF Winter Conference on
  Applications of Computer Vision}, 2023, pp. 289--299.

\bibitem{nichol2021glide}
A.~Nichol, P.~Dhariwal, A.~Ramesh, P.~Shyam, P.~Mishkin, B.~McGrew,
  I.~Sutskever, and M.~Chen, ``Glide: Towards photorealistic image generation
  and editing with text-guided diffusion models,'' \emph{ArXiv}, p.
  abs/2112.10741, 2021.

\bibitem{saharia2022photorealistic}
C.~Saharia, W.~Chan, S.~Saxena, L.~Li, J.~Whang, E.~L. Denton, K.~Ghasemipour,
  R.~Gontijo~Lopes, B.~Karagol~Ayan, T.~Salimans \emph{et~al.},
  ``Photorealistic text-to-image diffusion models with deep language
  understanding,'' \emph{Advances in Neural Information Processing Systems},
  vol.~35, pp. 36\,479--36\,494, 2022.

\bibitem{feng2023ernie}
Z.~Feng, Z.~Zhang, X.~Yu, Y.~Fang, L.~Li, X.~Chen, Y.~Lu, J.~Liu, W.~Yin,
  S.~Feng \emph{et~al.}, ``Ernie-vilg 2.0: Improving text-to-image diffusion
  model with knowledge-enhanced mixture-of-denoising-experts,'' in
  \emph{IEEE/CVF Conference on Computer Vision and Pattern Recognition}, 2023,
  pp. 10\,135--10\,145.

\bibitem{rombach2022high}
R.~Rombach, A.~Blattmann, D.~Lorenz, P.~Esser, and B.~Ommer, ``High-resolution
  image synthesis with latent diffusion models,'' in \emph{IEEE/CVF Conference
  on Computer Vision and Pattern Recognition}, 2022, pp. 10\,684--10\,695.

\bibitem{lugmayr2022repaint}
A.~Lugmayr, M.~Danelljan, A.~Romero, F.~Yu, R.~Timofte, and L.~Van~Gool,
  ``Repaint: Inpainting using denoising diffusion probabilistic models,'' in
  \emph{IEEE/CVF Conference on Computer Vision and Pattern Recognition}, 2022,
  pp. 11\,461--11\,471.

\bibitem{zhao2022egsde}
M.~Zhao, F.~Bao, C.~Li, and J.~Zhu, ``Egsde: Unpaired image-to-image
  translation via energy-guided stochastic differential equations,''
  \emph{Advances in Neural Information Processing Systems}, vol.~35, pp.
  3609--3623, 2022.

\bibitem{saharia2022image}
C.~Saharia, J.~Ho, W.~Chan, T.~Salimans, D.~J. Fleet, and M.~Norouzi, ``Image
  super-resolution via iterative refinement,'' \emph{IEEE Transactions on
  Pattern Analysis and Machine Intelligence}, vol.~45, no.~4, pp. 4713--4726,
  2022.

\bibitem{xiang2023denoising}
W.~Xiang, H.~Yang, D.~Huang, and Y.~Wang, ``Denoising diffusion autoencoders
  are unified self-supervised learners,'' \emph{ArXiv}, p. abs/2303.09769,
  2023.

\bibitem{baranchuk2021label}
D.~Baranchuk, I.~Rubachev, A.~Voynov, V.~Khrulkov, and A.~Babenko,
  ``Label-efficient semantic segmentation with diffusion models,''
  \emph{ArXiv}, p. abs/2112.03126, 2021.

\bibitem{sigdiff}
T.~Amit, T.~Shaharbany, E.~Nachmani, and L.~Wolf, ``Segdiff: Image segmentation
  with diffusion probabilistic models,'' \emph{ArXiv}, p. abs/2112.00390, 2021.

\bibitem{medsegdiff}
J.~Wu, H.~Fang, Y.~Zhang, Y.~Yang, and Y.~Xu, ``Medsegdiff: Medical image
  segmentation with diffusion probabilistic model,'' \emph{ArXiv}, p.
  abs/2211.00611, 2022.

\bibitem{medsegdiff2}
J.~Wu, R.~Fu, H.~Fang, Y.~Zhang, and Y.~Xu, ``Medsegdiff-v2: Diffusion based
  medical image segmentation with transformer,'' \emph{ArXiv}, p.
  abs/2301.11798, 2023.

\bibitem{ma2023diffusionseg}
C.~Ma, Y.~Yang, C.~Ju, F.~Zhang, J.~Liu, Y.~Wang, Y.~Zhang, and Y.~Wang,
  ``Diffusionseg: Adapting diffusion towards unsupervised object discovery,''
  \emph{ArXiv}, p. abs/2303.09813, 2023.

\bibitem{chen2023generative}
J.~Chen, J.~Lu, X.~Zhu, and L.~Zhang, ``Generative semantic segmentation,'' in
  \emph{IEEE/CVF Conference on Computer Vision and Pattern Recognition}, 2023,
  pp. 7111--7120.

\bibitem{le2023maskdiff}
M.-Q. Le, T.~V. Nguyen, T.-N. Le, T.-T. Do, M.~N. Do, and M.-T. Tran,
  ``Maskdiff: Modeling mask distribution with diffusion probabilistic model for
  few-shot instance segmentation,'' \emph{ArXiv}, p. abs/2303.05105, 2023.

\bibitem{UNet}
O.~Ronneberger, P.~Fischer, and T.~Brox, ``U-net: Convolutional networks for
  biomedical image segmentation,'' in \emph{International Conference on Medical
  Image Computing and Computer-Assisted Intervention}, 2015, pp. 234--241.

\bibitem{IDDPM}
A.~Q. Nichol and P.~Dhariwal, ``Improved denoising diffusion probabilistic
  models,'' in \emph{International Conference on Machine Learning}, 2021, pp.
  8162--8171.

\bibitem{U2Net}
X.~Qin, Z.~Zhang, C.~Huang, M.~Dehghan, O.~R. Zaiane, and M.~Jagersand,
  ``U2-net: Going deeper with nested u-structure for salient object
  detection,'' \emph{Pattern Recognition}, vol. 106, p. 107404, 2020.

\bibitem{swin}
Z.~Liu, Y.~Lin, Y.~Cao, H.~Hu, Y.~Wei, Z.~Zhang, S.~Lin, and B.~Guo, ``Swin
  transformer: Hierarchical vision transformer using shifted windows,'' in
  \emph{IEEE/CVF International Conference on Computer Vision}, 2021, pp.
  10\,012--10\,022.

\bibitem{vaswani2017attention}
A.~Vaswani, N.~Shazeer, N.~Parmar, J.~Uszkoreit, L.~Jones, A.~N. Gomez,
  {\L}.~Kaiser, and I.~Polosukhin, ``Attention is all you need,'' in
  \emph{Neural Information Processing Systems}, 2017, pp. 6000--6010.

\bibitem{he2016deep}
K.~He, X.~Zhang, S.~Ren, and J.~Sun, ``Deep residual learning for image
  recognition,'' in \emph{IEEE/CVF Conference on Computer Vision and Pattern
  Recognition}, 2016, pp. 770--778.

\bibitem{zhang2017beyond}
K.~Zhang, W.~Zuo, Y.~Chen, D.~Meng, and L.~Zhang, ``Beyond a gaussian denoiser:
  Residual learning of deep cnn for image denoising,'' \emph{IEEE transactions
  on image processing}, vol.~26, no.~7, pp. 3142--3155, 2017.

\bibitem{MSLSTIPT}
Y.~Sun, J.~Yang, and W.~An, ``Infrared dim and small target detection via
  multiple subspace learning and spatial-temporal patch-tensor model,''
  \emph{IEEE Transactions on Geoscience and Remote Sensing}, vol.~59, no.~5,
  pp. 3737--3752, 2020.

\bibitem{PSTNN}
L.~Zhang and Z.~Peng, ``Infrared small target detection based on partial sum of
  the tensor nuclear norm,'' \emph{Remote Sensing}, vol.~11, no.~4, p. 382,
  2019.

\bibitem{IPI}
C.~Gao, D.~Meng, Y.~Yang, Y.~Wang, X.~Zhou, and A.~G. Hauptmann, ``Infrared
  patch-image model for small target detection in a single image,'' \emph{IEEE
  transactions on image processing}, vol.~22, no.~12, pp. 4996--5009, 2013.

\bibitem{CIMD}
A.~Rahman, J.~M.~J. Valanarasu, I.~Hacihaliloglu, and V.~M. Patel, ``Ambiguous
  medical image segmentation using diffusion models,'' in \emph{IEEE/CVF
  Conference on Computer Vision and Pattern Recognition}, 2023, pp.
  11\,536--11\,546.

\bibitem{loshchilov2018decoupled}
I.~Loshchilov and F.~Hutter, ``Decoupled weight decay regularization,'' in
  \emph{International Conference on Learning Representations}, 2018.

\bibitem{warfield2004simultaneous}
S.~K. Warfield, K.~H. Zou, and W.~M. Wells, ``Simultaneous truth and
  performance level estimation (staple): an algorithm for the validation of
  image segmentation,'' \emph{IEEE transactions on medical imaging}, vol.~23,
  no.~7, pp. 903--921, 2004.

\end{thebibliography}

\vspace{11pt}

	\begin{IEEEbiography}[{\includegraphics[width=1in,height=1.25in,clip,keepaspectratio]{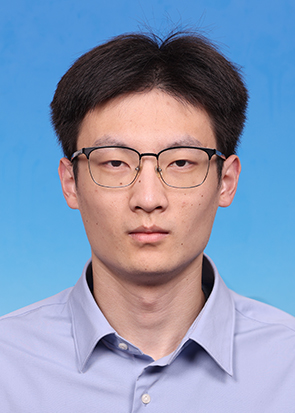}}]{Haoqing Li}
	received his B.Sc. degree in automation from Shandong University of Science and Technology, Qingdao, China, in 2021. He is currently pursuing an M.Sc. degree with the Department of Control Science and Engineering, Beijing University of Technology, Beijing. His research interests include computer vision, image semantic segmentation, deep learning, and their applications in infrared image processing.
\end{IEEEbiography}

\begin{IEEEbiography}[{\includegraphics[width=1in,height=1.25in,clip,keepaspectratio]{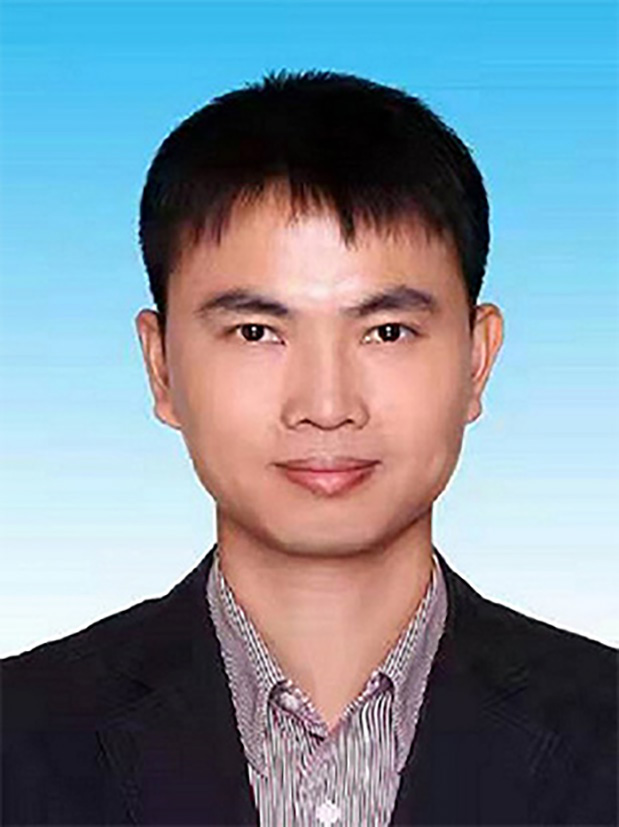}}]{Jinfu Yang}
	received his Ph.D. in pattern recognition and intelligent systems from the National Laboratory of Pattern Recognition, Chinese Academy of Sciences, in 2006. In 2013 - 2014, he was a visiting scholar at the University of Waterloo, Canada. Since 2006, he has been with the Faculty of Information Technology and the Beijing Key Laboratory of Computational Intelligence and Intelligent System, Beijing University of Technology, Beijing, where he is currently a Professor of computer vision. He is the secretary general of the Beijing Association for Artificial Intelligence, and a member of the Technical Committee of Computer Vision and Intelligent Robot of the Chinese Computer Federation (CCF CV and CCF TCIR). His research interests include pattern recognition, computer vision, and robot navigation.
\end{IEEEbiography}

\begin{IEEEbiography}[{\includegraphics[width=1in,height=1.25in,clip,keepaspectratio]{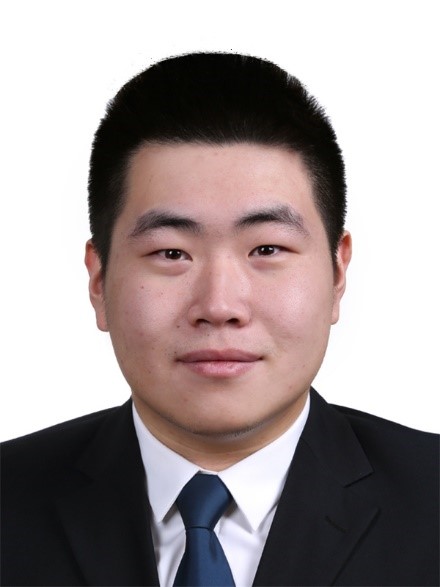}}]{Runshi Wang}
	received his B.Sc. degree in automation from Beijing University of Technology, in 2021. He is currently pursuing an M.Sc. degree with the Department of Control Science and Engineering, Beijing University of Technology, Beijing. His current research interests include small target detection, image-to-image translation computer vision, and robot environment perception.
\end{IEEEbiography}

\begin{IEEEbiography}[{\includegraphics[width=1in,height=1.25in,clip,keepaspectratio]{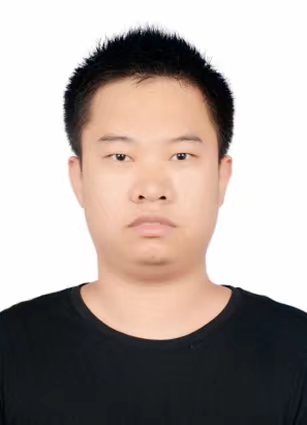}}]{Yifei Xu}
	received his B.Sc. degree in Measurement and control technology and instrument from Tianjin Polytechnic University, in 2021. He is currently pursuing an M.Sc. degree with the Department of Control Science and Engineering, Beijing University of Technology, Beijing. His research interests include deep neural network compression, computer vision, and robotic environmental awareness.
\end{IEEEbiography}

\vspace{11pt}

\vfill

\end{document}